\documentclass[letterpaper]{article} 
\usepackage{aaai2026}  
\usepackage{times}  
\usepackage{helvet}  
\usepackage{courier}  
\usepackage[hyphens]{url}  
\usepackage{graphicx} 
\usepackage{natbib}  
\usepackage{caption} 
\usepackage{algorithm}
\usepackage{algorithmic}

\usepackage{newfloat}
\usepackage{listings}
\DeclareCaptionStyle{ruled}{labelfont=normalfont,labelsep=colon,strut=off} 
\floatstyle{ruled}
\newfloat{listing}{tb}{lst}{}
\floatname{listing}{Listing}
\newtheorem{definition}{Definition}

\newtheorem{theorem}{Theorem}

\newtheorem{corollary}{Corollary}

\title{Incremental LTL$_f$ Synthesis}
\author {
    Giuseppe De Giacomo\textsuperscript{\rm 1,2},
    Yves Lespérance\textsuperscript{\rm 3},
    Gianmarco Parretti\textsuperscript{\rm 2},
    Fabio Patrizi\textsuperscript{\rm 2},
    Moshe Y. Vardi\textsuperscript{\rm 4}
}
\affiliations {
    \textsuperscript{\rm 1} University of Oxford\\
    \textsuperscript{\rm 2} University of Rome ``La Sapienza''\\
    \textsuperscript{\rm 3} York University\\
    \textsuperscript{\rm 4} Rice University\\
    giuseppe.degiacomo@cs.ox.ac.uk, lesperan@yorku.ca, \{parretti, patrizi\}@diag.uniroma1.it, vardi@cs.rice.edu
}

\usepackage{bibentry}

\usepackage{xspace}

\newcommand{\myi}{(\emph{i})\xspace}
\newcommand{\myii}{(\emph{ii})\xspace}
\newcommand{\myiii}{(\emph{iii})\xspace}
\newcommand{\myiv}{(\emph{iv})\xspace}

\newcommand{\exptime}{\textsc{exptime}\xspace}
\newcommand{\twoexptime}{2\textsc{exptime}\xspace}
\newcommand{\threeexptime}
{3\textsc{exptime}\xspace}

\newcommand{\LTLf}{\textsc{ltl}$_f$\xspace}

\newcommand{\LTL}{\textsc{ltl}\xspace}

\newcommand{\act}{Act}

\newcommand{\NNF}{\textsc{nnf}\xspace}

\newcommand{\DFA}{\textsc{dfa}\xspace}
\newcommand{\DFAs}{\textsc{dfa}s\xspace}
\newcommand{\Play}{\mathit{Play}}
\newcommand{\run}{\mathit{Run}}

\newcommand{\FOND}{\textsc{fond}\xspace}
\newcommand{\PDDL}{\textsc{pddl}\xspace}

\newcommand{\BDD}{\textsc{bdd}\xspace}

\newcommand{\cudd}{\textsc{cudd}\xspace}

\newcommand{\lydia}{\textsc{lydia}\xspace}
\newcommand{\mona}{\textsc{mona}\xspace}
\newcommand{\syft}{\textsc{syft}\xspace}
\newcommand{\isabel}{\textsc{isabel}\xspace}
\newcommand{\isabeldp}{\textsc{isabel}-\textsc{dp}\xspace}
\newcommand{\isabelfp}{\textsc{isabel}-\textsc{fp}\xspace}

\newcommand{\plants}{\textsc{plants}\xspace}
\newcommand{\requests}{\textsc{requests}\xspace}
\newcommand{\counter}{\textsc{counter}\xspace}
\newcommand{\tireworld}{\textsc{tireworld}\xspace}

\newcommand{\Math}[1]{\ensuremath{#1}}

\newcommand{\modecal}[1]{{\Math{\mathcal{#1}}}}

\newcommand{\A}{\modecal{A}} 
 
\newcommand{\E}{\modecal{E}} 
\newcommand{\G}{\modecal{G}} 
 
 \renewcommand{\L}{\modecal{L}}

\newcommand{\U}{\modecal{U}} 
 \newcommand{\X}{\modecal{X}}
\newcommand{\Y}{\modecal{Y}} 
\newcommand{\R}{\modecal{R}}

\newcommand{\mnext}{\raisebox{-0.27ex}{\LARGE$\circ$}}	
\newcommand{\malways}{\square}		
\newcommand{\meventually}{\lozenge}	
\newcommand{\muntil}{\mathop{\U}}	
\newcommand{\mrelease}{\mathop{\R}}	

\newcommand{\mweaknext}{\mdblkcircle}
\newcommand{\mdblkcircle}{\raisebox{-0.27ex}{\LARGE$\bullet$}}

\newcommand{\limp}{\supset}

\usepackage{amssymb}
\usepackage{paralist}
\usepackage{multirow}
\usepackage{amsmath}
\usepackage{subcaption}

\begin{document}

\maketitle

\begin{abstract}
    In this paper, we study incremental \LTLf synthesis -- a form of reactive synthesis where the goals are given incrementally while in execution.  
    In other words, the protagonist agent is already executing a strategy for a certain goal when it receives a new goal: at this point, the agent has to abandon the current strategy and synthesize a new strategy still fulfilling the original goal, which was given at the beginning, as well as the new goal, starting from the current instant.  In this paper, we formally define the problem of incremental synthesis and study its solution. We propose a solution technique that efficiently performs incremental synthesis for multiple \LTLf goals by leveraging auxiliary data structures constructed during automata-based synthesis. 
%
%
We also consider an alternative solution technique based on \LTLf formula progression. We show that, in spite of the fact that formula progression can generate formulas that are exponentially larger than the original ones, their minimal automata remain bounded in size by that of the original formula. On the other hand, we show experimentally that, if implemented naively, i.e., by actually computing the automaton of the progressed
\LTLf
formulas
from scratch every time a new goal arrives, the solution based on formula progression is not competitive. 
\end{abstract}
\section{Introduction}
In this paper, we study \emph{incremental} \LTLf synthesis -- a form of reactive synthesis where the goals are given incrementally while in execution.
Reactive synthesis was originally studied in Formal Methods for Linear Temporal Logic on infinite traces (\LTL)~\cite{pnueli1977temporal,PnueliR89} and uses techniques drawn from model checking. More recently,  synthesis has been studied for Linear Temporal Logic on finite traces \cite{DegVa13}, where the agent must eventually stop after completing its goal \cite{DegVa15}. \LTLf synthesis has some nice computational characteristics that allow one to exploit symbolic techniques typical of model checking \cite{BaierK08}, to obtain symbolic solvers significantly more scalable than those for \LTL, see \cite{ZhuTLPV17,bansal2020hybrid,DeGiacomoF21,KankariyaB24,ZhuF25,DuretLutzZPDV25}. 


In AI, reactive synthesis is essentially a variant of strong planning for temporally extended goals in Fully Observable Nondeterministic (\FOND) domains \cite{Cimatti03,BacchusK-AIJ00,BaierFM07,DeGiacomoR18,CamachoM-IJCAI19,DeGiacomoPZ23}. 

Incremental synthesis deals with the case where an agent is already executing a strategy for a certain goal when it gets an additional new goal: at this point, the agent has to abandon the strategy for the original goal that it is executing and synthesize a new strategy that fulfills both the original and new goal. Crucially,  the original goal has to be fulfilled starting from the beginning, through the current history, while the new goal starts from the current instant.
In general, incremental synthesis may be needed several times during the execution. 

Incremental synthesis shares some similarities with compositional approaches to synthesis, where the \LTLf specification is a conjunction of subformulas that are handled separately \cite{bansal2020hybrid,BansalGSLVZ22}. Here, however, the conjunction involves formulas to be evaluated over traces that start at different time points. 
Incremental synthesis also shares some similarities with live synthesis, where the synthesis is performed after a certain history and not in the initial state \cite{FinkbeinerKM22,ZhuD-KR22}.

Obviously, when adopting a new goal, we must guarantee the realizability of the new goal together with the previous ones.
In case of conflict, we need to decide whether to forfeit the adoption of the new goal or drop some of the old ones to keep the adopted goals realizable. While we do not discuss how to resolve this conflict in the paper, leaving it to future work, our techniques do indeed assess whether the conflict is present.



In this paper, we formally define the problem of incremental synthesis and study its solution.  In particular, we propose a solution technique that leverages auxiliary data structures constructed during automata-based synthesis. 
We show its soundness, completeness, and use it to characterize the worst-case computational complexity of the incremental \LTLf synthesis as \twoexptime-complete -- as for standard \LTLf synthesis. 
We then evaluate its effectiveness experimentally using a prototype implementation. 

We also study an alternative way of solving incremental synthesis by  exploiting \LTLf formula progression, i.e., the fixpoint characterization of temporal formulas \cite{GabbayPSS80,Manna1982,Emerson1990TemporalAM,BacchusK-AIJ00,DeGiacomoFLVXZ22}, which allows to recursively split an \LTLf formula into a propositional formula on the \emph{current} instant and a temporal formula to be checked at the \emph{next} instant.
By applying formula progression, we can reduce the problem of incremental synthesis to standard synthesis for the conjunction of the progressed original goal and the new one. Progressed formulas can become exponentially larger than the initial ones, so one could think that incremental synthesis using this technique is \threeexptime. Instead, here we show that the automata of progressed formulas (once minimized) are bounded by the automata of their corresponding original formulas. Hence, also using \LTLf progression, incremental synthesis can be solved in \twoexptime.  
However, the \LTLf progression-based technique, if implemented naively, i.e., by actually computing the automata of the progressed formulas every time a new goal arrives, is not really competitive, as we show empirically in Section~\ref{sec:experiments}. In fact, the automata-based solution technique we present here can be seen as a clever implementation of the solution based on formula progression which takes advantage of automata caching.

\section{Preliminaries~\label{sec:pre}}

We consider specifications (aka \emph{formulas}) in Linear Temporal Logic on \emph{finite traces} (\LTLf)~\cite{DegVa13}. \LTLf is a formalism widely used in AI to specify temporal properties over finite-time horizons, e.g., in planning, temporally extended goals~\cite{CamachoTMBM17,DeGiacomoR18,CamachoM-IJCAI19} as well as state/action trajectory constraints~\cite{BonassiGPS21,BonassiGS22}.

We require \LTLf specifications to be in Negation Normal Form (\NNF):  negation can appear in front of atomic propositions only. This does not cause any loss of generality: every \LTLf formula can be transformed into \NNF in linear time. Formally, given a set of \emph{atomic propositions} (aka \emph{atoms}) $P$, the syntax of \LTLf formulas $\varphi$ in \NNF is:

\smallskip
{\small \centerline{$
\varphi := p \mid \lnot p \mid \varphi_1 \land \varphi_2 \mid \varphi_1 \lor \varphi_2 \mid \mnext \varphi \mid \mweaknext \varphi \mid \varphi_1 \muntil \varphi_2 \mid \varphi_1 \mrelease \varphi_2
$}}
\smallskip

\noindent Where $p \in P$. Boolean operators include: \emph{negation} ($\lnot$); \emph{disjunction} ($\lor$); and \emph{conjunction} ($\land$). Temporal operators include: \emph{next} ($\mnext$); \emph{weak next} ($\mweaknext$); \emph{until} ($\muntil$); and \emph{release} ($\mrelease$). 
Additional operators include: standard operators of propositional logic, such as $true$, $false$, \emph{implication} ($\limp$), and \emph{equivalence} ($\equiv$); and temporal operators defined as abbreviations, such as \emph{eventually} ($\meventually \varphi \doteq true \muntil \varphi$), and \emph{always} ($\malways \varphi \doteq false \mrelease \varphi$).
The size of $\varphi$, denoted $|\varphi|$, is the number of subformulas in its abstract syntax tree. 

\LTLf formulas are interpreted over finite traces $\tau\in (2^P)^*$  
of propositional interpretations over $P$. The empty trace is $\epsilon$.  We denote by $\tau_i \in 2^{P}$ the propositional interpretation in the $i$-th time step of $\tau$ and by $|\tau|$ its length. Given the traces $\tau = \tau_0 \cdots \tau_n$ and $\tau' = \tau'_0 \cdots \tau'_m$, their concatenation is the trace $\tau \cdot \tau' = \tau_0 \cdots \tau_n \cdot \tau'_0 \cdots \tau'_m$.
Given an \LTLf formula $\varphi$, a trace $\tau$, and an index $i$ such that $0\leq i< |\tau|$,
the following inductive definition formalizes when
$\tau$ \emph{satisfies} $\varphi$ at $i$:
\begin{compactitem}
    \item $\tau, i \models p$ iff $p \in \tau_i$, for $p \in P$;
    \item $\tau, i \models \lnot p$ iff $p \not \in \tau_i$, for $p \in P$;
    \item $\tau, i \models \varphi_1 \lor \varphi_2$ iff $\tau, i \models \varphi_1$ or $\tau, i \models \varphi_2$;
    \item $\tau, i \models \varphi_1 \land \varphi_2$ iff $\tau, i \models \varphi_1$ and $\tau, i \models \varphi_2$;
    \item $\tau, i \models \mnext \varphi$ iff $i < |\tau|-1 $ and $\tau, i + 1 \models \varphi$;
    \item $\tau, i \models \mweaknext \varphi$ iff $i < |\tau|-1 $ implies $\tau, i + 1 \models \varphi$;
    \item $\tau, i \models \varphi_1 \muntil \varphi_2$ iff there exists $j$ such that $i \leq j < |\tau|$ and $\tau, j \models \varphi_2$, and, for every $k$ such that $i \leq k < j$ we have $\tau, k \models \varphi_1$;
    \item $\tau, i \models \varphi_1 \mrelease \varphi_2$ iff either: (1) there exists $j$ such that $i \leq j < |\tau|$ and $\tau, j \models \varphi_1$ and, for every $k$ such that $i \leq k \leq j$, we have $\tau, k \models \varphi_2$; \emph{or} (2) for every $k$ such that $i \leq k < |\tau|$, we have $\tau, k \models \varphi_2$.
\end{compactitem}

A finite trace $\tau$ satisfies $\varphi$, denoted $\tau \models \varphi$, if $\tau, 0 \models \varphi$. 




\emph{\LTLf (reactive) synthesis} concerns finding a strategy to satisfy an \LTLf goal specification. Goals are expressed as \LTLf formulas over $P = \Y \cup \X$, where $\Y$ and $\X$ are disjoint sets of atoms under the control of the agent and environment respectively. Traces over $\Y \cup \X$ are denoted $\tau = (Y_0 \cup X_0)(Y_1 \cup X_1) \cdots$, where $Y_i \in 2^\Y$ and $X_i \in 2^\X$ for every $i \geq 0$. Infinite traces of this form are called \emph{plays}; finite traces are also called \emph{histories}. 


An \emph{(agent) strategy} is a function $\sigma: (2^\X)^* \rightarrow 2^\Y$ that maps (possibly empty) sequences of environment moves to an agent move. The domain of $\sigma$ includes the empty trace $\epsilon$ as we assume that the agent moves first -- as in planning.
A play \mbox{$\tau = (Y_0 \cup X_0)(Y_1 \cup X_1) \cdots$} is $\sigma$-consistent if: \myi $Y_0 = \sigma(\epsilon)$; and \myii  $Y_{i} = \sigma(X_0 \cdots X_{i-1})$ for every $i > 0$. 
That a history $h$ is $\sigma$-consistent is defined analogously. The set of $\sigma$-consistent plays is denoted $\Play(\sigma)$.  

A strategy $\sigma$ is winning for $\varphi$ if, for every play $\tau \in Play(\sigma)$, there exists a finite prefix $\tau^k$ that satisfies $\varphi$, i.e., $\tau^k \models \varphi$. 
\LTLf synthesis is the problem of finding a winning strategy for $\varphi$, if any exists. If there exists a winning strategy for $\varphi$, we say that synthesis is \emph{realizable}; otherwise, it is \emph{unrealizable}. \LTLf synthesis is \twoexptime-complete wrt~$\varphi$~\cite{DegVa15}.

\section{Incremental Synthesis\label{sec:incrsynth}}









In this paper, we study \emph{incremental \LTLf synthesis}. In incremental synthesis, the agent does not know the goals in advance -- as in standard synthesis -- but receives them online, during execution. 


Specifically, consider an agent that is executing a strategy $\sigma$ -- previously synthesized for a goal $\varphi_{org}$ -- which has so far generated a history $h$. 
Assume that at the end of $h$ a new goal $\varphi_{new}$ arrives. On the one hand, the agent cannot rely on the original strategy $\sigma$, as it was synthesized for $\varphi_{org}$, without even knowing $\varphi_{new}$;  on the other hand, the agent cannot simply synthesize 
a new strategy $\sigma'$ for both $\varphi_{org}$ and $\varphi_{new}$, since  
$\sigma'$ would restart satisfying $\varphi_{org}$ from the current instant, without taking into account that after the history $h$ only part of the original goal $\varphi_{org}$ remains to be realized.
Instead, we want $\sigma'$ to: (R1) guarantee the satisfaction of $\varphi_{new}$; and (R2) continue with the satisfaction of $\varphi_{org}$ from where $\sigma$ left, i.e., taking into account the history $h$. We formalize this intuition below. 

Requirement (R1) can be easily captured by requiring the new strategy to be winning for $\varphi_{new}$. Regarding requirement (R2): we formalize a new notion that captures what it means for a strategy $\sigma$ to be winning for $\varphi$ taking into account that a history $h$ has occurred. That is, $\sigma$ executed at $h$ only generates plays that satisfy $\varphi$ from the beginning. Formally:

\begin{definition}\label{def:win-assuming-h}
    Let $\varphi$ be an \LTLf goal and $h$ a history. A strategy $\sigma$ is winning for $\varphi$ \emph{assuming} $h$ if, for every $\tau \in Play(\sigma)$, we have that $\tau' = h\cdot \tau$ has a finite prefix that satisfies $\varphi$.
\end{definition}


Having notions that capture requirements (R1) and (R2), we can formally define the incremental synthesis problem:

\begin{definition}\label{def:incrsynth}
Let: $\varphi_{org}$ be an original \LTLf goal; $h$ a history; and $\varphi_{new}$ a new goal. Incremental synthesis is the problem of computing a strategy $\sigma$ such that: \begin{compactenum}
    \item $\sigma$ is winning for $\varphi_{org}$ assuming $h$; and
    \item $\sigma$ is winning for $\varphi_{new}$;
\end{compactenum}
if any exists. In this latter case, we say that incremental synthesis is \emph{realizable}; otherwise, it is \emph{unrealizable}.
\end{definition}


Incremental synthesis generalizes standard synthesis. Indeed, we have the following: 
synthesis for $\varphi$ is equivalent to incremental synthesis for $\varphi_{org} = \varphi$, $h = \epsilon$, and $\varphi_{new} = true$ (or alternatively for $\varphi_{org} = true$, $h = \epsilon$, and $\varphi_{new} = \varphi_{org}$). Given that standard synthesis is \twoexptime-hard~\cite{DegVa15}, we obtain the hardness of incremental synthesis:

\begin{theorem}\label{thm:incr-synth-hardness}
    Incremental synthesis is \twoexptime-hard wrt $\varphi_{org}$ and $\varphi_{new}$.
\end{theorem}

In fact, incremental synthesis also preserves the realizability for the original goal. Say that $\sigma_{org}$ is a winning strategy for $\varphi_{org}$ that the agent was using when $h$ occurred and $\varphi_{new}$ arrived; also, say that $\sigma_{new}$ is a strategy that satisfies Definition~\ref{def:incrsynth}. Consider the overall strategy obtained by combining the executions of $\sigma_{org}$ and $\sigma_{new}$: we have that this strategy is winning for $\varphi_{org}$ -- i.e., the \emph{realizability of $\varphi_{org}$ is preserved}. The intuition is as follows: before $h$ occurred, the agent was using $\sigma_{org}$, that always guarantees the satisfaction of $\varphi_{org}$; after $h$ occurred, the agent uses $\sigma_{new}$, that guarantees the satisfaction of $\varphi_{org}$ in every extension of $h$. Formally: denote by $\sigma_{org}[h \leftarrow \sigma_{new}]$ the strategy that agrees with $\sigma_{org}$ everywhere, except in $h$ and all its extensions, where it agrees with $\sigma_{new}$. 
Then we have:
\begin{theorem}\label{thm:std-incr-relation}
    The strategy $\sigma_{org}[h \leftarrow \sigma_{new}]$, where $\sigma_{org}$ is winning for $\varphi_{org}$ and consistent with $h$, and $\sigma_{new}$ is a strategy that satisfies Definition~\ref{def:incrsynth}, is winning for $\varphi_{org}$.
\end{theorem}

We conclude this section by observing that Definition~\ref{def:incrsynth} can be easily generalized to an arbitrary number of goals. 
For instance, consider the original goal $\varphi_{org}$, and two new successive goals $\varphi_{new_1}$, arriving at history $h_1$ and $\varphi_{new_2}$ arriving at history $h_1 \cdot h_2$. Incremental synthesis in this case consists computing a strategy that is:
\myi winning for $\varphi_{org}$ assuming $h_1 \cdot h_2$; \myii winning for $\varphi_{new_1}$ assuming $h_2$; and \myiii winning for $\varphi_{new_2}$.



\smallskip

\section{Automata-Based Synthesis Technique\label{sec:direct}}

In this section, we present an automata-based technique to solve incremental synthesis. The technique is based on solving reachability games played over suitable deterministic finite automata, which we briefly review below.

A \emph{deterministic finite automaton} (\DFA) is a tuple $\A=(\Sigma,S,\iota,\delta,F)$, where: $\Sigma$ is a finite input alphabet; $S$ is a finite set of states; $\iota \in S$ is the initial state; $\delta: S \times \Sigma \rightarrow S$ is the transition function; and $F \subseteq S$ is the set of final states. 
The size of $\A$ is $|S|$. 
For convenience, we extend $\delta$ to a function $\delta: S \times \Sigma^* \rightarrow S$ over finite traces $\alpha = \alpha_0 \cdots \alpha_n$, inductively defined as follows: 
\myi $\delta(s, \epsilon) = s$; and \myii
$\delta(s, \alpha_0 \cdots \alpha_{n}) = 
\delta(\delta(s, \alpha_0 \cdots \alpha_{n-1}), \alpha_{n})$. A trace $\alpha$ is \emph{accepted} if \mbox{$\delta(\iota, \alpha) \in F$}. The \emph{language} of $\A$, denoted $\L(\A)$, is the set of traces that $\A$ accepts. A \DFA $\A$ is \emph{minimal} if there is no other \DFA $\A'$ such that $\L(\A) = \L(\A')$ and $|\A'| < |\A|$. A \DFA can be minimized in polynomial time by using an algorithm that repeatedly collapses into one states that are indistinguishable wrt to all possible future inputs. For the languages that we consider in this paper, called \emph{regular languages}, there exists a unique minimal \DFA~\cite{Sipser97}. 

\begin{theorem}~\cite{DegVa15}\label{thm:ltlf2dfa}
Given an \LTLf formula $\varphi$, we can construct a \DFA $\A_{\varphi}$ of size at most double exponential in that of $\varphi$ and whose language is the set of finite traces that satisfy $\varphi$.    
\end{theorem}

We note, however, that the worst-case doubly-exponential blowup while constructing \DFAs of \LTLf formulas is rare in practice, and the construction scales well, see, e.g.,~\cite{DegVa15,ZhuTLPV17,bansal2020hybrid,DeGiacomoF21}. 


For synthesis, we see such a \DFA  as a game. Specifically, a \DFA game is a \DFA $\G = (\Sigma, S,\iota, \delta, F)$
with input alphabet $\Sigma = 2^{\Y \cup \X}$, 
where $\Y$ and $\X$ are disjoint sets of \emph{atoms} under control of agent and environment, respectively. 
A \emph{game strategy} is a function $\kappa: S \rightarrow 2^{\Y}$ that maps game states to agent moves. The notion of play in Section~\ref{sec:pre} also applies to \DFA games. A play $\tau = (Y_0 \cup X_0)(Y_1 \cup X_1) \cdots$ is $\kappa$-consistent if: \myi $Y_0 = \kappa(\iota)$; and \myii  $Y_i = \kappa(\delta(\iota, \tau^{i-1}))$ for every $i > 0$, where $\tau^{i-1} = (Y_0 \cup X_0) \cdots (Y_{i-1} \cup X_{i-1})$. 
The set of $\kappa$-consistent plays is $\Play(\kappa)$. A game strategy $\kappa$ is winning in $\G$ if, for every play $\tau \in \Play(\kappa)$, there exists a finite prefix $\tau^k$ that is accepted by $\G$. That is: in a \DFA game $\G$ the agent has to visit at least once the set of final states -- thus generating only traces accepted by $\G$. The winning region $W$ is the set of states $s$ where the agent has a winning strategy in the game $\G' = (\Sigma, S, s, \delta, F)$, i.e., the same game as $\G$, but with new initial state $s$. There exists a winning game strategy in $\G$ iff $\iota \in W$. Solving a \DFA game $\G$ is the problem of computing the winning region $W$ and a winning game strategy $\kappa$, if any exists. \DFA games can be solved in polynomial time by a backward-induction algorithm that performs a least fixpoint computation over the state space of the game~\cite{AG11}. Given a game $\G$ and a winning game strategy $\kappa: S  \rightarrow 2^{\Y}$, we can construct in polynomial time a winning strategy $\sigma_{(\G, \kappa)}: (2^{\X})^* \rightarrow 2^{\Y}$ for the corresponding \LTLf formula as follows:
\myi $\sigma_{(\G, \kappa)}(\epsilon) = \kappa(\iota)$; and \myii for every $\tau = (Y_0 \cup X_0) \cdots (Y_n \cup X_n)$, define $\sigma_{(\G, \kappa)}(X_0 \cdots X_n) = \kappa(\delta(\iota, \tau))$. 
The pair $(\G, \kappa)$ is a \emph{transducer} that implements the strategy $\sigma_{(\G, \kappa)}$.

\smallskip

Now, we present the algorithm for incremental synthesis. Intuitively, the algorithm consists of four steps that do the following: Step (1) constructs the \DFAs that accept the traces that satisfy the original and new goal; Steps (2) and (3) manipulate and combine these \DFAs so to obtain a \DFA game that admits a winning strategy iff incremental synthesis is realizable; Step~(4) solves this game to decide whether incremental synthesis is realizable -- in which case it returns a strategy to satisfy 
both the original and new goal. 

The algorithm uses two polynomial operators to manipulate \DFAs, \emph{product} and \emph{progression}: the former combines several \DFAs into one whose language is the intersection of the languages of its operands; the latter progresses the initial state of a \DFA through a history $h$, so that the progressed \DFA accepts exactly the traces accepted by the input \DFA and that contain $h$ as a prefix. Formally: 

\begin{definition}[\DFA Product]\label{def:product}
    Let $\A_1, \cdots, \A_n$ be \DFAs where $\A_i = (\Sigma, S_i, \iota_i, \delta_i, F_i)$ for every $i$. Their product is the \DFA $\A = \A_1 \times \cdots \times \A_n = (\Sigma, S, \iota, \delta, F)$, with: alphabet $\Sigma$; state space $S = S_1 \times \cdots \times S_n$; initial state $\iota = (\iota_1, \cdots, \iota_n)$; transition function $\delta((s_1, \cdots, s_n), w) = (\delta(s_1, w), \cdots, \delta(s_n, w))$ for every $w \in \Sigma$; final states $F = F_1 \times \cdots \times F_n$; and language $\L(\A) = \bigcap_i \L(\A_i)$.
\end{definition}

\begin{definition}[\DFA Progression]\label{def:dfa-progression}
    Let $\A = (\Sigma, S, \iota, \delta, F)$ be a \DFA and $h$ a history. The \DFA progression of $\A$ through $h$ is \DFA $\A_{h} = (\Sigma, S, \delta(\iota, h), \delta, F)$ -- i.e., the same as $\A$, but with initial state $\delta(\iota, h)$, so that its language is $\L(\A_{h}) = \{h \cdot \tau ~|~\tau \in \L(\A)\}$. 
\end{definition}


We give below the algorithm for incremental synthesis.


\smallskip

\noindent \textbf{Algorithm~1.} Consider an instance of incremental synthesis: $\varphi_{org}$ is the original \LTLf goal; $h$ the history; and $\varphi_{new}$ the new \LTLf goal. \smallskip \begin{compactenum}
    \item[(1)] Construct the \DFAs $\A_{org}$ and $\A_{new}$ that accept the traces that satisfy $\varphi_{org}$ and $\varphi_{new}$, respectively; 
    \item[(2)]  Construct the \DFA $\A_{(org, h)}$ as the \DFA progression of $\A_{org}$ through $h$; 
    \item[(3)] Construct the \DFA game $\G = \A_{(org, h)} \times \A_{new}$. Say $\iota$ is the initial state of $\G$; 
    \item[(4)] Solve $\G$. Let $W$ be the winning region and $\kappa$ a winning game strategy, if any: if $\iota \not \in W$, return that incremental synthesis is \emph{unrealizable}; else, return the strategy $\sigma_{(\G, \kappa)}$ as the transducer $(\G, \kappa)$. 
\end{compactenum}


\smallskip


We show below the correctness of Algorithm 1.
Recall that we are interested in constructing a game that admits a winning strategy iff incremental synthesis is realizable. 

The \DFA game is obtained as $\G = {\A_{(org, h)} \times \A_{new}}$; by Definition~\ref{def:product}, the traces that $\G$ accepts are those accepted by both $\A_{(org, h)}$ and $\A_{new}$. As a result, $\G$ accepts the traces $\tau$ such that: (P1) $\tau \models \varphi_{new}$, by Theorem~\ref{thm:ltlf2dfa}; (P2) $h \cdot \tau \models \varphi_{org}$, by Theorem~\ref{thm:ltlf2dfa} and Definition~\ref{def:dfa-progression}.

Next, we show that $\G$ admits a winning strategy iff incremental synthesis is realizable. Assume that $\G$ admits a winning strategy $\sigma_{(\G, \kappa)}$: by (P1) and (P2), we have that $\sigma_{(\G, \kappa)}$ generates only plays $\tau$ such that $\tau$ has a finite prefix that satisfies $\varphi_{new}$ and $\tau' = h \cdot \tau$ has a finite prefix that satisfies $\varphi_{org}$ -- i.e., $\sigma_{(\G, \kappa)}$ satisfies Definition~\ref{def:incrsynth} and incremental synthesis is realizable. Conversely, assume that $\G$ does not admit a winning strategy: by (P1) and (P2), there is no strategy that satisfies Definition~\ref{def:incrsynth} -- i.e., incremental synthesis is unrealizable. As a result, we proved that Algorithm~1 is sound and complete, which is formally established in the following:

\begin{theorem}\label{thm:alg1-correct}
    Incremental synthesis is realizable iff Algorithm~1 returns a strategy. 
\end{theorem}

Regarding complexity: Algorithm~1 is polynomial in the length of $h$, as it updates the initial state of $\A_{org}$ by executing its transition function through all of $h$. Wrt $\varphi_{org}$ and $\varphi_{new}$: Algorithm~1 solves a game that is worst-case doubly-exponential in their sizes; solving this game can be done in polynomial time -- so that Algorithm~1 establishes membership of incremental synthesis in \twoexptime wrt $\varphi_{org}$ and $\varphi_{new}$. Having established the hardness of incremental synthesis in Theorem~\ref{thm:incr-synth-hardness}, we obtained the computational complexity of incremental synthesis:
\begin{theorem}\label{thm:incrsynthcomplexity}
    Incremental synthesis for $\varphi_{org}$, $h$, and $\varphi_{new}$ is: \begin{compactitem}
        \item \twoexptime-complete wrt $\varphi_{org}$ and $\varphi_{new}$;
        \item Polynomial wrt $h$.
    \end{compactitem}
\end{theorem}


Finally, we show Algorithm~1 can be extended to an arbitrary number of goals. Let $\varphi_{org}$ be the original goal, $h = h_1 \cdot \ldots \cdot h_{n}$ a history, and $\varphi_{i}$ is a new goal that arrives at the end of $h_1 \cdot \ldots \cdot h_i$ for every $i \leq n$. The key modifications to Algorithm~1 are: (1) construct the \DFAs $\A_{org}$ that accepts the traces that satisfy $\varphi_{org}$, and $\A_{i}$ that accepts the traces that satisfy $\varphi_i$ for every $i \leq n$; (2) compute the \DFA progressions $\A_{(org, h_1 \cdot \ldots \cdot h_n)}$, and $\A_{(i, h_{i+1} \cdot \ldots \cdot h_n)}$ for every $i < n$ (recall the \DFA for the last new goal $\varphi_n$ is not progressed); (3) construct and solve \DFA game {$\G =\A_{(org, h_1 \cdot \ldots \cdot h_n)} \times \A_{(1, h_2 \cdot \ldots \cdot h_n)} \times \cdots \times \A_{(n-1, h_n)} \times \A_{n}$}. The arguments for soundness and completeness above can be extended by induction to the case where multiple goals are considered; the complexity wrt goals and histories remains unchanged.



Obviously, an agent that uses incremental synthesis does not need to compute from scratch the \DFAs of the original goals every time it is requested to add a new goal -- an operation that would cost \twoexptime for every goal. Instead, the agent can store in its memory the \DFAs of the original goal and progress their initial states at each time step during execution; when the agent is requested to add a new goal, it only needs to compute the \DFA for the new goal. Recall: \emph{the bottleneck of \LTLf synthesis is the \DFA construction}~\cite{ZhuTLPV17,bansal2020hybrid,DeGiacomoF21,DeGiacomoFLVXZ22}; hence, storing the \DFA of the goals in memory has the potential to improve significantly the performance of incremental synthesis.

These features make Algorithm~1 well-suited for efficient implementation, as confirmed empirically in Section~\ref{sec:experiments}.

\section{Formula Progression\label{sec:progression}}

In this section, we show that incremental synthesis can be solved by
\emph{\LTLf formula progression} (aka \emph{\emph{\LTLf progression}})~\cite{DeGiacomoFLVXZ22,GabbayPSS80,Manna1982,Emerson1990TemporalAM,BacchusK-AIJ00}. Intuitively, the progression of an \LTLf formula is a syntactic rewriting that captures, given a history, what remains to be satisfied of that formula after the history has occurred. 

We start by reviewing \LTLf progression. First: \LTLf is interpreted over finite traces; it is necessary to clarify when traces end. To do so: we introduce two formulas $\malways(false)$ and $\meventually(true)$, which, intuitively, denote \emph{finite trace ends} and \emph{finite trace does not end}, respectively. Let $\varphi$ be an \LTLf formula in \NNF over a set of atoms $P$ and $w \in 2^P$ a propositional interpretation. The progression of $\varphi$ through $w$, denoted $prog(\varphi, w)$, is~\cite{DeGiacomoFLVXZ22}:
\[
{\begin{array}{l}
    prog(p,w) \doteq true \mbox{ if } p \in w \mbox{ and } false \mbox{ otherwise}\\
    prog(\lnot p, w) \doteq true \mbox{ if } p \not \in w \mbox { and } false \mbox { otherwise} \\ 
    prog(\varphi_1 \lor\varphi_2,w) \doteq prog(\varphi_1,w) \lor prog(\varphi_2,w)\\
    prog(\varphi_1 \land \varphi_2,w) \doteq prog(\varphi_1,w) \land prog(\varphi_2,w) \\
    prog(\mnext \varphi,w) \doteq \varphi \land \meventually(true) \\ prog(\mweaknext \varphi, w) \doteq \varphi \lor \malways(false) \\
    prog(\varphi_1 \muntil \varphi_2,w) \doteq \\
    ~~~~prog(\varphi_2,w) \lor (prog(\varphi_1,w) \land prog(\mnext(\varphi_1 \muntil \varphi_2), w) \\
    prog(\varphi_1 \mrelease \varphi_2,w) \doteq \\
    ~~~~prog(\varphi_2,w) \land (prog(\varphi_1,w) \lor prog(\mweaknext(\varphi_1 \mrelease \varphi_2), w) 
\end{array}}
\]

Let $h = \alpha_0 \cdots \alpha_n \in (2^P)^*$ be a history. We extend the \LTLf progression of $\varphi$ to $h$, denoted $prog(\varphi, h)$, as follows: \myi $prog(\varphi, \epsilon) = \varphi$; and \myii $prog(\varphi, \alpha_0 \cdots \alpha_n) = prog(prog(\varphi, \alpha_0 \cdots \alpha_{n-1}), \alpha_n)$. Intuitively, the \LTLf progression of $\varphi$ through $h$ captures what remains to be satisfied of $\varphi$ after $h$ has occurred. 


Consider an instance of incremental synthesis: $\varphi_{org}$ is the original \LTLf goal; $h$ is the history; and $\varphi_{new}$ is the new \LTLf goal. 
Using \LTLf progression, we can solve incremental synthesis by standard synthesis: 

\begin{theorem}\label{thm:incr-to-std}
    A strategy $\sigma$ solves
    incremental synthesis for $\varphi_{org}$, $h$, and $\varphi_{new}$ iff $\sigma$ solves standard synthesis for $prog(\varphi_{org}, h) \land \varphi_{new}$.
\end{theorem}






Next, we investigate the complexity of solving incremental synthesis by \LTLf progression. We observe that progressing $\varphi_{org}$ through $h$ generates an \LTLf formula $prog(\varphi_{org}, h)$ that is worst-case exponential in the size of $\varphi_{org}$~\cite{DeGiacomoFLVXZ22}; this observation, together with the complexity of \LTLf synthesis, may hint that the complexity of solving incremental synthesis by \LTLf progression is \threeexptime wrt $\varphi_{org}$ and \twoexptime wrt $\varphi_{new}$. However, \emph{the bound wrt $\varphi_{org}$ is too pessimistic}. In fact, we show below that the size of the \DFA of any progressed formula is indeed bounded by that of the \DFA of the original formula~-- so that also solving incremental synthesis by \LTLf progression establishes \twoexptime membership wrt $\varphi_{org}$. 






To show this, we analyze the relation between \LTLf progression and \DFA progression, see Definition~\ref{def:dfa-progression}. Let $\A_{\varphi}$ be the \DFA for an \LTLf goal $\varphi$; say that $\iota$ and $\delta$ are initial state and transition function of $\A_{\varphi}$, respectively. If we progress $\varphi$ one step through $w$, this corresponds to performing that same step in $\A_{\varphi}$ as well: the corresponding \DFA is $\A_{\varphi}$, but with its initial state progressed to $\delta(\iota, w)$. This result can be extended by induction to histories of arbitrary length: if we progress $\varphi$ through $h$, its corresponding \DFA is $\A_{(\varphi, h)}$ -- i.e., the same as $\A_{\varphi}$, but with its initial state progressed to $\delta(\iota, h)$, as in Definition~\ref{def:dfa-progression}. Hence, we obtain:

\begin{theorem}\label{thm:ltlf-dfa-progression}
    Let: $\varphi$ be an \LTLf goal; $\A_{\varphi}$ its corresponding \DFA; and $h$ an history. The \DFA progression $\A_{(\varphi,h)}$ of $\A_\varphi$ through $h$ accepts exactly the traces that satisfy $prog(\varphi, h)$.
\end{theorem}

\begin{figure}[t]
    \centering
    \includegraphics[width=0.8\linewidth]{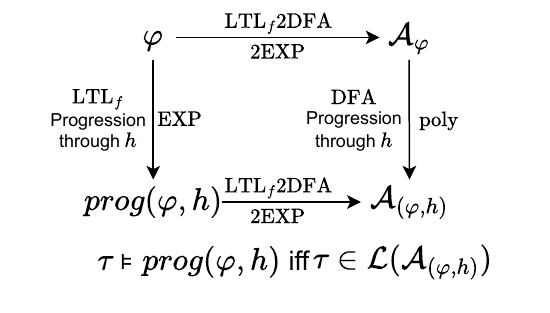}
    \caption{Relation between \LTLf progression and \DFA progression through a history $h$}
    \label{fig:ltlf-dfa-prog}
\end{figure}

Theorem~\ref{thm:ltlf-dfa-progression} shows that while $prog(\varphi, h)$ is worst-case exponential in the size of $\varphi$,
its corresponding \DFA can be obtained in \twoexptime~-- vs \threeexptime~-- as the \DFA progression $\A_{(\varphi,h)}$ of $\A_{\varphi}$ through $h$. In other words: the worst-case exponential blowup introduced by the \LTLf progression $prog(\varphi, h)$ is not reflected in its corresponding minimal \DFA, which is essentially the same as $\A_{\varphi}$. This result is summarized in Figure~\ref{fig:ltlf-dfa-prog}.


\smallskip
In fact, progressing the initial state of $\A_{\varphi}$ to get $\A_{(\varphi, h)}$ may cause parts of $\A_{\varphi}$ to be unreachable, which gives opportunities for further minimization. As a result: 

\begin{theorem}\label{thm:dfa-progression-size}
    Let $\A_{\varphi}$ be the \DFA for an \LTLf formula $\varphi$ and $h$ a history. We have that $|\A_{(\varphi,h)}| \leq |\A_{\varphi}|$. 
\end{theorem}

Finally, we show that Algorithm~1 can be seen as a clever implementation of incremental synthesis by \LTLf progression. Indeed, this follows by Theorems~\ref{thm:incr-synth-hardness},~\ref{thm:incr-to-std}, and~\ref{thm:ltlf-dfa-progression}, as well as Definitions~\ref{def:product} and~\ref{def:dfa-progression}:

\begin{corollary}\label{cor:incr-std-smart}
    Consider an instance of incremental synthesis: $\varphi_{org}$ is the original \LTLf goal; $h$ the history; $\varphi_{new}$ the new \LTLf goal. The \DFA $\G = \A_{(org, h)} \times \A_{new}$ constructed (in polynomial time) in Step (3) of Algorithm~1 accepts exactly the traces that satisfy \mbox{$prog(\varphi_{org}, h) \land \varphi_{new}$}.
\end{corollary}

\section{Implementation}\label{sec:implementation}

We implemented the automata-based technique (Algorithm~1) extended to multiple goals in a tool called \isabeldp\footnote{\url{https://github.com/GianmarcoDIAG/ISabel/}} (\emph{\textbf{I}ncremental \textbf{S}ynthesis by \textsc{\textbf{d}fa} \textbf{P}rogression}). \isabeldp employs the symbolic synthesis framework in~\cite{ZhuTLPV17} -- which is integrated into state-of-the-art \LTLf synthesis tools~\cite{ZhuF25,DuretLutzZPDV25}. We use \lydia~\cite{DeGiacomoF21}, which is among the best-performing \LTLf-to-\DFA conversion tools, to generate \DFAs from \LTLf formulas; in turn, \lydia uses \mona~\cite{HenriksenJJKPRS95} to represent \DFAs and perform relevant \DFA manipulations, such as, e.g., product and minimization. \mona uses a semi-symbolic representation of \DFAs: states are represented explicitly; transitions are represented symbolically. We use \syft to represent \DFAs fully symbolically and solve symbolic \DFA games~\cite{ZhuTLPV17}. \syft represents state space and transition function of symbolic \DFAs using Binary Decision Diagrams~\cite{Bryant92}, with \cudd-3.0.0~\cite{Somenzi} as the underlying \BDD library.

Our implementation employs both \mona's and \syft's representations of \DFAs in order to gain the maximal benefit during incremental synthesis. Specifically: (1) The \mona \DFA representation is employed during game construction and \DFA progression, so as to gain the maximal benefit from \DFA minimization; (2)  The \syft \DFA representation is employed during game solving, so as to gain the maximal benefit from symbolic game-resolution techniques.

Following the approach outlined in Section~\ref{sec:direct}, \isabeldp maintains a data structure that stores the \mona \DFAs of the added goals in a set $V$; their initial states are progressed incrementally during execution. When a new goal $\varphi_{new}$ arrives, the implementation retrieves the \DFAs of the previously added goals from $V$; doing so enables us to minimize the number of \LTLf-to-\DFA conversions -- thus saving \twoexptime for each stored goal. Specifically, the implementation takes the following steps: (1) transform $\varphi_{new}$ into the \mona \DFA $\A_{new}$; (2) construct the minimal \mona \DFA game $\G$ as the product of the \DFAs in $V$ and $\A_{new}$; (3) transform $\G$ into a \syft symbolic \DFA game $\G^s$; (4) solve $\G^s$ and return a new strategy iff incremental synthesis is realizable, in which case $\A_{new}$ is also added to $V$. 

 
\section{Empirical Analysis}\label{sec:experiments}
In this section, we present an empirical analysis for incremental synthesis. The goal of the analysis is to establish whether incremental synthesis is feasible in settings where standard synthesis is. Moreover, we want to determine whether the automata-based solution is more effective than the solution based on formula progression, when implemented directly.

To do so, we compare the performance of \isabeldp to that of a \emph{baseline} implementation that solves incremental synthesis by \LTLf progression on some scalable benchmarks. The latter implementation is called \isabelfp (\emph{\textbf{I}ncremental \textbf{S}ynthesis by \textsc{ltl}$_f$ \textbf{F}ormula \textbf{P}rogression}) and, for a fair comparison, it also uses \lydia for \LTLf-to-\DFA conversion and \syft for symbolic game-resolution.
We will show that \emph{\isabeldp performs consistently and significantly better than \isabelfp~--~thus showing the effectiveness of the automata-based approach}.




\paragraph{Experimental Methodology.} We considered four \emph{benchmarks}: two of our own invention, \plants and \requests; two adapted from existing benchmarks used in \LTLf synthesis and strong Fully Observable Nondeterministic (\FOND) planning, \counter and \tireworld.\footnote{In many AI applications -- including planning -- the agent has a model of the world in which it operates, e.g., a \FOND domain.
World models are typically considered specifications of the environment and can often be expressed in \LTLf, e.g., for the case of a \FOND domain, see~\cite{AminofDMR19}. To avoid the introduction of explicit environment specifications, we exploit the following: once the agent action has been performed and the environment reaction observed, e.g., the effect of a \PDDL \texttt{one-of} condition has occurred, the fluents describing the successor domain state are \emph{completely determined}~\cite{DeGiacomoPZ23}. It follows that we can assign fluents as agent atoms and allow the environment reaction to be completely free (possibly with some care in handling all environment reactions appropriately). As a result, we can represent the world model as part of the original agent goal $\varphi_{org}$. This is the approach we follow in \tireworld benchmarks. 
}  
These benchmarks have been selected as they enable the agent to add many goals during execution -- which makes them suited for evaluating incremental synthesis implementations. In each benchmark, an \emph{instance} consists of the following: the implementation is given up to 3600 seconds until it times out; we evaluate the number of goals added by both 
\isabel implementations before the timeout, as well as the time required to add each goal. To reproduce the incremental setting, the addition of new goals is interleaved with the execution of the strategy synthesized for the previous goals: at each time step, the agent performs an action and adds a new goal. All experiments were run on a laptop with an operating system 64-bits Ubuntu 22.04, 3.6 GHz CPU, and 8 GBs memory.

\paragraph{Benchmarks.} We give a description for each benchmark. We provide the full specification for \tireworld and \plants benchmarks; we refer to the supplementary material for the full specification of \requests and \counter. Recall from the synthesis settings in Sections~\ref{sec:pre} and~\ref{sec:incrsynth}: atoms are divided into agent atoms $\Y$ and environment atoms $\X$.

In our \tireworld benchmarks~\cite{MuiseMB12}, the agent must navigate between several locations dealing with nondeterministic outcomes while moving. Specifically: from location $\ell$
the agent can move to any location $\ell'$; as the agent moves, one its tires can nondeterministically become flat (in $\ell'$); when a tire is flat, the agent cannot move, but can change the flat tire (we assume that spare tires are always available). Formally, \tireworld benchmarks are modeled as follows. The agent atoms are: \myi $at_{i}$, denotes that the current location of the agent is $i$; \myii $move_{i}$, denotes that the agent moves to location $i$; and \myiii $change\text{-}tire$, denotes that the agent changes a flat tire. There is only one environment atom: \myiv $make\text{-}flat$, denotes that the environment caused one of the agent tires to become flat. Instances in \tireworld benchmarks are parametrized by the number of locations $\ell$. We assume that the environment can make a tire flat only when the agent moves (\emph{not} when the agent changes a flat tire), written in \LTLf $\alpha \doteq \malways (make\text{-}flat \limp \bigvee_{i < \ell} move_i)$. In each instance, we generate goals $\varphi_{n}$, with $n \leq 30$, as follows: if the assumption $\alpha$ holds, the agent must traverse locations from $0$ to $\ell-1$ in sequence, for a total of $2n + 1$ visits, written in \LTLf $\varphi_{n} \doteq$
\centerline{\small $\alpha \limp (\meventually(at_0 \land \mnext (\meventually at_1 \land \cdots \land \mnext( \meventually at_{\ell-1} \land$}
\centerline {\small $~~~~~~~~~~~~~~~~~~~~\mnext (\meventually at_0 \land \cdots \land \mnext (\meventually at_{2n+1~mod~\ell}) \cdots)))))$}

\noindent
where $n$ is the goal parameter.
The original goal ($n = 0$) also requires the agent to satisfy the specification of the domain in which it operates -- which in this case is a \FOND planning domain where nondeterminism is captured by the environment atom $make\text{-}flat$. The domain specification includes \textbf{(C1)} \emph{Initial state}: the agent is at location $\ell_0$, written:

\centerline{\small $\varphi_{init} \doteq at_0 \land \bigwedge_{i.0 < i < \ell} \lnot at_{i}$}

\noindent \textbf{(C2)} \emph{Action preconditions}: at each time step, the agent can move only if it does not have a flat tire, and can change a tire only if it has a flat tire, written $\varphi_{pre} \doteq $

\centerline{\small$\malways(\bigwedge_{i<\ell} (move_{i} \limp \lnot flat\text{-}tire)) \land \malways(change\text{-}tire \limp flat\text{-}tire)$} 

\noindent \textbf{(C3)} \emph{Mutual exclusion axiom}: at each time step, the agent must perform exactly one action, written

\centerline{\small $\varphi_{mutex} \doteq \malways(change\text{-}tire \lor \bigvee_{i<\ell} move_i) \land$}
\centerline{\small $\malways (change\text{-}tire \limp \lnot \bigvee_{i < \ell} move_{i}) \land$}
\centerline{\small $\malways(\bigwedge_{i<\ell} move_{i} \limp \lnot change\text{-}tire \land \lnot \bigvee_{j \neq i} move_{j})$}

\noindent \textbf{(C4)} \emph{Transition function}: at each time step, the agent must follow the transition function of the domain, written

\centerline{\small $\varphi_{trans} \doteq \malways(\bigwedge_{i \leq \ell} \bigwedge_{j \neq i} at_{i} \land move_{j} \land make\text{-}flat \limp$} 
\centerline{\small $\mweaknext (at_{j} \land flat\text{-}tire \land \bigwedge_{k \neq j} \lnot at_{k}))~\land$}
\centerline{\small $\malways(\bigwedge_{i \leq \ell} \bigwedge_{j \neq i} at_{i} \land move_{j} \land \lnot make\text{-}flat \limp$} 
\centerline{\small $\mweaknext (at_{j} \land \lnot flat\text{-}tire \land \bigwedge_{k \neq j} \lnot at_{k}))~\land$}
\centerline{\small $\malways(\bigwedge_{i \leq \ell} at_{i} \land change\text{-}tire \limp$}
\centerline {\small $\mweaknext(at_{i} \land \lnot flat\text{-}tire \land \bigwedge_{j \neq i} \lnot at_{j}))$}

As a result, the original goal is:

\centerline{\small $\varphi_{org} \doteq \varphi_{init} \land \varphi_{pre} \land \varphi_{mutex} \land \varphi_{trans} \land \varphi_{0}$}

For every goal, a winning strategy for the agent is the following: navigate locations $0$ to $\ell-1$ in order to accomplish a total of $2n+1$ visits; if the environment makes a tire flat, change the tire and resume navigation. We consider \tireworld instances with varying numbers of locations $\ell$ such that $1 \leq \ell \leq 10$, denoted \tireworld-$\ell$; in each instance, we consider goals with $n \leq 30$, for a total of $300$ goals.

We remark that there are more effective ways to generate the \DFA of a \FOND domain than converting its \LTLf specification into \DFA. For instance,~\cite{DeGiacomoDP25} show a \FOND-to-\DFA conversion technique that does not incur in the \twoexptime blowup resulting from the \LTLf-to-\DFA conversion, but only costs \exptime in the size of the domain.
Nonetheless, we represent the domain specification as an \LTLf formula in order to be consistent with the synthesis settings introduced in Sections~\ref{sec:pre} and~\ref{sec:incrsynth}.



In \counter benchmarks~\cite{ZhuDPV20}, the specification is as follows: the agent maintains a $k$-bits counter, with all bits initially set to 0; at each time step, the environment chooses whether to issue an increment request for the counter, captured by the environment atom $add$; the agent chooses whether to accept the request, and then increment the counter. Instances in \counter benchmarks are parametrized by the number of bits $k$ in the counter. 
In each instance, we generate goals $\varphi_{n}$ as follows: if the environment always issues increment requests, the agent must eventually set the counter value to $2n + 1$, where $n$ is the goal parameter. One such goal is written in \LTLf as $\varphi_{n} \doteq (\malways add) \limp \meventually \varphi^{counter}_{2n + 1}$, where $\varphi^{counter}_{2n + 1}$ is a conjunction of agent atoms that sets the counter value to $2n + 1$. 
The original goal also requires the agent to satisfy the counter specification.
For every goal, a winning strategy is to accept all increment requests and increment the counter accordingly. We consider \counter instances with number of bits $k$ s.t. $6 \leq k \leq 10$, denoted \counter-$k$; in each instance, \mbox{we consider goals with $n \leq 30$, for a total of $150$ goals.} 

In \plants benchmarks, the specification is as follows: the agent has to take care of several plants; the agent can water the plants, captured by the agent atom $water$; that a plant $i$ is alive is captured by the environment atom $alive_i$; the environment can also rain, captured by the environment atom $rain$. Instances in \plants benchmarks are parametrized by the number of plants $p$. 
At every time step. a plant $i$ will be alive iff $i$ was alive in the previous time step and either the agent waters it or the environment rains, written:

\centerline{\small $\alpha_i \doteq \malways(\mweaknext(alive_i) \equiv alive_i \land (water \lor rain))$}

In each instance, we generate goals as follows: if the assumption $\alpha_i$ holds, the agent must take care of plant $i$ for $3(n+1)$ days, each day being a distinct time step, where $n$ is the goal parameter. One such a goal is  written 

\centerline{\small $\varphi_n \doteq \bigwedge_{i < p} \alpha_i \limp \varphi_{3(n+1)}^{alive_i}$ with $\varphi_{3(n+1)}^{alive_i} \doteq \meventually(\bigwedge_{j < 3(n+1)} \mnext^j alive_i$)}

\noindent where $\mnext^k$ denotes $k$ nested $\mnext$ operators. 
For every goal, a winning strategy is to always water the plants. We consider \plants instances with number of plants $p$ s.t. $1 \leq p \leq 10$, denoted \plants-$p$; in each instance, we consider goals with $n \leq 30$, for a total of $300$ goals.

In \requests benchmarks, the specification is as follows: the agent provides a set of services that the environment, at each time step, can request;
to satisfy a request, the agent must perform a specific sequence of actions; 
the environment issues a finite number of consecutive requests, and the agent must satisfy all of them. Instances in \requests benchmarks are parametrized by: (1) the number $i$ of services $s_i$ the agent provides; and (2) the number $j$ of actions $a_{i,j}$ that the agent must perform to provide service $i$ in response to a request for it. 
In each instance, we generate goals as follows: the agent must satisfy $2n + 1$ environment requests, where $n$ is the goal parameter. For every goal, a winning strategy is to perform all actions to satisfy the requests issued by the environment. We consider \requests instances with: \myi number of services $i$ s.t. $1 \leq i \leq 3$; \myii number of actions $j$ s.t. $1 \leq j \leq 4$; such instances are denoted \requests-$i$-$j$; in each instance, we consider goals with $n \leq 40$, for a total of $480$ goals.

 \begin{table}[t!]
     \centering
     \resizebox{0.80\linewidth}{!}{
     \begin{tabular}{|c||c c|c c|}
          \hline
          \multirow{2}{4em}{Instance} & \multicolumn{2}{c|}{Added Goals} & \multicolumn{2}{c|}{Avg. Runtime (s)}\\
          & \isabeldp & \isabelfp & \isabeldp & \isabelfp \\
          \hline\hline
\plants-1 & \textbf{30}/30 & 19/30 & \textbf{9.12} & 155.12 \\
\plants-2 & \textbf{25}/30 & 16/30 & \textbf{15.68} & 172.00 \\
\plants-3 & \textbf{22}/30 & 14/30 & \textbf{18.70}  & 179.43\\
\plants-4 & \textbf{20}/30 & 13/30 & \textbf{19.60}  & 173.17 \\
\plants-5 & \textbf{19}/30 & 12/30 & \textbf{29.94} & 265.45 \\
\plants-6 & \textbf{18}/30 & 12/30 & \textbf{24.85} & 216.06 \\
\plants-7 & \textbf{17}/30 & 11/30 & \textbf{35.37}  & 268.26\\
\plants-8 & \textbf{16}/30 & 11/30 & \textbf{35.37} & 268.26 \\
\plants-9 & \textbf{15}/30 & 10/30 & \textbf{32.49}  & 190.29 \\
\plants-10 & \textbf{15}/30 & 10/30 & \textbf{42.09}  & 237.39 \\
\counter-6 & \textbf{25}/30 & 20/30 & \textbf{25.02}  & 29.00 \\
\counter-7 & \textbf{21}/30 & 12/30 & \textbf{3.98}  & 13.18 \\
\counter-8 & \textbf{19}/30 & 10/30 & \textbf{14.57}   & 41.11\\
\counter-9 & \textbf{0}/30 & \textbf{0}/30 & - & - \\
\counter-10 & \textbf{0}/30 & \textbf{0}/30 & - & - \\
\requests-1-1 & \textbf{36}/40 & 28/40 & \textbf{27.53}  & 107.01 \\
\requests-1-2 & \textbf{36}/40 & 29/40 & \textbf{31.91} & 117.05 \\
\requests-1-3 & \textbf{36}/40 & 29/40 & \textbf{31.41} & 115.76 \\
\requests-1-4 & \textbf{36}/40 & 29/40 & \textbf{31.80} & 111.20\\
\requests-2-1 & \textbf{37}/40 & 29/40 & \textbf{24.39}  & 112.03 \\
\requests-2-2 & \textbf{37}/40 & 29/40 & \textbf{23.83}  & 112.27 \\
\requests-2-3 & \textbf{36}/40 & 29/40 & \textbf{24.49}  & 105.88 \\
\requests-2-4 & \textbf{36}/40 & 29/40 & \textbf{27.97}  & 105.64 \\
\requests-3-1 & \textbf{37}/40 & 29/40 & \textbf{24.12}  & 113.95 \\
\requests-3-2 & \textbf{36}/40 & 29/40 & \textbf{24.85} & 108.55 \\
\requests-3-3 & \textbf{35}/40 & 29/40 & \textbf{33.81}  & 111.29 \\
\requests-3-4 &  \textbf{6}/40 &  \textbf{6}/40 & 7.09  & \textbf{6.43}\\
\tireworld-1 & \textbf{30}/30 & 26/30  & \textbf{5.52} & 118.14 \\          
\tireworld-2 & \textbf{30}/30 & 26/30  & \textbf{5.42} & 121.49 \\
\tireworld-3 & \textbf{30}/30 & 25/30 & \textbf{5.06} & 141.56 \\
\tireworld-4 & \textbf{30}/30 & 24/30 & \textbf{9.73} & 117.57 \\
\tireworld-5 & \textbf{28}/30 & 21/30 & \textbf{17.89} & 152.42 \\
\tireworld-6 & \textbf{21}/30 & 19/30 & \textbf{32.04} & 151.25 \\
\tireworld-7 & \textbf{17}/30 & \textbf{17}/30 & \textbf{41.77} & 157.20 \\
\tireworld-8 & \textbf{15}/30 & \textbf{15}/30 & \textbf{34.05} & 98.96 \\
\tireworld-9 & \textbf{13}/30 & \textbf{13}/30 & \textbf{18.95}  & 47.23 \\
\tireworld-10 & \textbf{12}/30 & \textbf{12}/30 & \textbf{15.97}  & 36.65 \\
\hline \hline
\textbf{Total} & \textbf{892}/1230 & 692/1230 & & \\
\hline
     \end{tabular}}
     \caption{Comparison of the number of goals added and average time to add a goal achieved by \isabeldp and \isabelfp. Results of the best-performing implementation in bold.}
     \label{tab:comparison}
 \end{table}


 \paragraph{Empirical Results.} Table~\ref{tab:comparison} compares the number of goals added by \isabeldp and \isabelfp, as well as their average runtime to add a new goal; for a fair comparison, in each instance, the average times are computed wrt the goals that are added by both. First, we note that both \isabeldp and \isabelfp failed to add any goal in \counter-$9$ and \counter-$10$: this is because both timed out during the construction of the \DFA of the counter specification -- whose size grows exponentially in the number of bits. We have an analogous result in \requests-$3$-$4$: both \isabeldp and \isabelfp timed out during the construction of the \DFA of the $7$-th goal. These results confirm that \DFA construction is the bottleneck of incremental synthesis, as in standard synthesis.

 In the remaining instances, we have the following: \myi in terms of number of added goals, \isabeldp performs consistently better or the same as \isabelfp; \myii in terms of average time to add a new goal, \isabeldp is significantly better than \isabelfp~-- \emph{up to gaining one or two orders of magnitudes better performance in average}. This is evident in \plants and \requests: in their instances, the size of the progressed \LTLf goals may grow exponentially. To add a new goal, \isabelfp employs an \LTLf-to-\DFA conversion for the \LTLf formula obtained as the conjunction of the progressed \LTLf goals -- which may then cause an unnecessary overhead; on the other hand, \isabeldp employs an \LTLf-to-\DFA conversion for the new \LTLf goal only, and generates the \DFAs of the progressed \LTLf goals in polynomial time using \DFA progression -- which prevents any unnecessary overhead to occur.

 We have analogous results in \counter instances. However, it is important to note the following: \emph{in \counter instances, the size of the progressed \LTLf goals is almost the same as that of the non-progressed goals}. Nonetheless, \isabeldp outperformed \isabelfp both in terms of added goals and average time to add a goal. The reason is as follows: \isabeldp benefits from retrieving the \DFAs of the goals from memory, whereas \isabelfp incurs an unnecessary overhead by employing an \LTLf-to-\DFA conversion for the \LTLf formula obtained as the conjunction of the progressed \LTLf goals. Hence, storing the \DFAs of the goals in memory improves the performance of incremental synthesis.  

Regarding \tireworld benchmarks, we have the following. (1) Instances \tireworld-$1$ to \tireworld-$6$ show that \isabeldp performs better than \isabelfp both in terms of added goals and average time to add a goal -- \emph{up to gaining one or two orders of magnitude better performance in average} -- since \isabeldp avoids the unnecessary overhead resulting by using \LTLf progression. (2) Instance \tireworld-$7$ to \tireworld-$10$ show that \isabeldp and \isabelfp add the same number of goals, and, while \isabeldp remains much faster than \isabeldp, the performance gap between \isabeldp and \isabelfp becomes smaller. This latter result is due to memory limitation (8 GBs of RAM) of the machine on which the experiments are run.
Indeed, as we approach the memory limit, memory management overhead becomes significant for both \isabeldp and \isabelfp, until \emph{they both run out of memory during the construction of the \DFA game arena for computing the new strategy}.
Recall that \isabeldp and \isabelfp construct the same \DFA game arena (see Theorem~\ref{thm:incr-to-std} and Corollary~\ref{cor:incr-std-smart}): hence, the observed performance difference stems from the more efficient \DFA game arena construction employed by \isabeldp~-- which, in turn, suggests \isabeldp would scale better than \isabelfp in large \tireworld instances when run on more powerful hardware. 

Overall, the empirical analysis shows the superiority of \isabeldp to \isabelfp ~-- which confirms the effectiveness of our approach to incremental synthesis and its feasibility in settings where also standard synthesis is feasible. (See also the supplementary material for further empirical results.)

\section{Conclusion}

In addition to being of theoretical and general interest, this work also has a direct practical impact, providing an effective approach for revising an agent's strategy when goals dynamically change. 
%
%
%
%
Our results provide an efficient incremental technique for revising strategies for autonomous agents that operate in nondeterministic domains, reducing as much as possible the synthesis effort required when goals change dynamically. In this context, the approach can be directly integrated into a Goal/Intention Management System such as the one sketched in~\cite{DeGiacomoAAMAS25}.

Observe that
a new goal might be inconsistent with the current ones, leading to a situation where no strategy exists that guarantees the realization of all the agent's goals. Although the present paper does not address this issue, our technique allows for effective conflict detection, 
a first step toward resolution.
Then, if all the agent's goals have the same importance, one can look into finding a \emph{maximal realizable subset}, though this is an exponential problem itself, since it may require trying exponentially many 
combinations, each of which is an incremental synthesis problem.
A simple alternative is to assume
that the adopted goals are totally ordered with respect to priority, and given this, keep the set of highest priority goals that are jointly realizable.

Note that our approach can express a broad range of temporally extended goals/constraints in \LTLf, which is important for safety critical applications where formal guarantees of safe/correct behavior are required \cite{FerrandoDCFAM21}.
Mainstream autonomous agent development methods based on BDI agent programming \cite{RaoAgentspeak96,JasonBook07}, including those that support goal reasoning such as \cite{HarlandMTY-JAAMAS14} do not ensure that the agent's goals/intentions remain consistent/realizable and do not provide such safety/correctness guarantees.

In future work, we plan to investigate 
methods for performing incremental synthesis based on
forward search techniques adopted in synthesis and planning~\cite{CamachoM-IJCAI19,BonassiDFFGS23,MuiseMB24,DuretLutzZPDV25}. 
Incremental synthesis can be extended in several directions, e.g., synthesizing best-effort/good-enough strategies when there is no winning strategy~\cite{AminofDR21,DeGiacomoPZ23,AlmagorK20} or when the environment dynamics change~\cite{AminofDPR24}, as well as the partial observability setting~\cite{TabajaraV20}. These also remain interesting topics for future work.

\section*{Acknowledgments}

This work has been supported by the UKRI Erlangen AI Hub on Mathematical and Computational Foundations of AI (No. EP/Y028872/1), the Italian National Ph.D. in Artificial Intelligence atSapienza University of Rome, the National Science and Engineering Research Council of Canada, and York University.

\bibliography{aaai2026}

\newpage 

\phantom{abc}

\newpage

\centerline {\textbf{\huge Supplementary Material}}

\smallskip


\noindent The content of the supplementary material is as follows: in Section~\ref{sec:specs}, we provide additional details regarding the benchmarks introduced in Section~\ref{sec:experiments}; in Section~\ref{sec:experiments_bis}, we provide additional empirical results, specific for each instance in our benchmarks, and further discussion.


\section{Additional Details on Benchmarks}\label{sec:specs}

First, we give the full specification of \counter and \requests benchmarks.

\smallskip

We recall the specification of \counter benchmarks: the agent maintains a $k$-bits counter, with all bits initially set to 0; at each time step, the environment chooses whether to issue an increment request for the counter; the agent chooses whether to accept the request, and then increment the counter. Instances in \counter benchmarks are parametrized by the number of bits $k$ in the counter. Formally, \counter benchmarks are modeled as follows. There is one environment atom: \myi $add$, that denotes that the environment issues an increment requests for the counter. The agent atoms include: \myii $k$ atoms $\{b_0, \cdots, b_{k-1}\}$, that denote the value of the counter bits; and \myiii $k+1$ atoms $\{c_0, \cdots, c_{k}\}$ that denote carry bits, where $c_0$ denotes that the agent granted an increment request. In each instance, we generate goals $\varphi_{n}$ as follows: if the environment always issues increment requests, the agent must eventually set the counter value to $2n + 1$, where $n$ is the goal parameter. One such goal is written in \LTLf as $\varphi_{n} \doteq (\malways add) \limp \meventually \varphi^{counter}_{2n + 1}$, where $\varphi^{counter}_{2n + 1}$ is a conjunction of the bit atoms that sets the counter value to $2n + 1$. The original goal ($n = 0$) also requires the agent to satisfy the counter specification -- which specifies how the agent should update the value of bits and carry bits. The counter specification includes: \textbf{(C1)} \emph{Initial state}: every bit and carry bit evaluates to $0$ in the initial time step, written:

\centerline{\small $\varphi_{init} \doteq \bigwedge_{i < k} \lnot b_{i} \land \bigwedge_{i \leq k} \lnot 
c_i$}

\noindent \textbf{(C2)} \emph{Action preconditions}: at each time step, the agent can accept an increment request only if the environment issued an increment request in the previous time step, written:

\centerline{\small $\varphi_{pre} \doteq \malways(\mweaknext c_0 \limp add)$}

\noindent \textbf{(C3)} \emph{Transition function}: at each time step, the agent must update the value of the bits according to the specification of the counter, written

\centerline{ \small $\varphi_{trans} \doteq \bigwedge_ {i < k}\malways(B_i)$}
\noindent Where, for every $i < k$,  we have that $B_i \doteq$

\centerline{\small $(\lnot c_i \land \lnot b_i) \limp (\mweaknext (\lnot b_i \land \lnot c_{i+1})) \land$}
\centerline{\small $(\lnot c_i \land b_i) \limp (\mweaknext (b_i \land \lnot c_{i+1})) \land$}
\centerline{\small $(c_i \land \lnot b_i) \limp (\mweaknext (b_i \land \lnot c_{i+1})) \land$}
\centerline{\small $(c_i \land b_i) \limp (\mweaknext (\lnot b_i \land  c_{i+1}))$}

As a result, the original goal is: 

\centerline{\small $\varphi_{org} \doteq \varphi_{init} \land \varphi_{pre} \land \varphi_{trans} \land \varphi_{0}$}

It should be noted the following: \emph{at each time step, the size of the \LTLf progression of the goals in \counter instances is almost the same as that of the non-progressed goals}. In fact, at each time step, the progression of the goals consists of: \myi a propositional formula to be evaluated in the current time step -- which will disappear by progressing to either $true$ or $false$ in the subsequent time step; and \myii a temporal formula that will be evaluated in the subsequent time step. As an example, observe that the progression of $\varphi_{n}$ for every $w$ has the following shape:

\centerline{\small $prog(\varphi_n, w) = (add \land prog(\mweaknext(\malways(add)), w))  \limp$}
\centerline{\small 
$~~~~~~~~~(\varphi^{counter}_{2n+1} \lor prog(\mnext(\meventually(\varphi^{counter}_{2n+1})), w))$}

\noindent Similarly applies to the progression of the original goal.

\smallskip

We recall the specification of \requests benchmarks: the agent provides a set of services that the environment, at each time step, can request;
to satisfy a request, the agent must perform a specific sequence of actions; 
the environment issues a finite number of consecutive requests, and the agent must satisfy all of them. Instances in \requests benchmarks are parametrized by: (1) the number $i$ of services $s_i$ the agent provides; and (2) the number $j$ of actions $a_{i,j}$ that the agent must perform to provide service $i$ in response to a request for it. Formally, \requests benchmarks are formalized as follows. There are: \myi $i$ environment atoms $r_i$, each denoting that the environment requested service $i$; and \myii $i \times j$ agent atoms $a_{i,j}$, each denoting the $j$-th action needed for the agent to satisfy a request for service $i$. We consider the following formulas: \textbf{(F1)} The environment issues a sequence of $k$ requests, written: 

\centerline{\small $\E_{k} \doteq (\bigwedge_{v < k} \mweaknext^v (\bigvee_{s < i} r_s)) \land \mweaknext^{k} (\malways (\lnot \bigvee_{s < i} r_s))$}

\noindent Where $\mweaknext^k$ denotes $k$ nested $\mweaknext$ operators.

\noindent \textbf{(F2)} For every type of request $i$, the agent must perform a specific sequence of $j$ actions, written: 

\centerline{\small $\varphi_{request} \doteq \bigwedge_{s < i} r_s \limp$} \centerline{\small $(\meventually(a_{s,1} \land \mnext (\meventually(a_{s,2}) \land \cdots \land \mnext(\meventually (a_{s,j})) \cdots)$}

\noindent \textbf{(F3)} The agent must generate a trace of length at least $k$ -- so that it reads all requests generated by the environment -- enforced by:

\centerline{\small $\varphi^{true}_{k} = \mnext^{k}(true)$}

\noindent Where $\mnext^k$ denotes $k$ nested $\mnext$ operators. In each request instance, we generate goals as follows: if the environment generates a sequence of $2n + 1$ requests, the agent must satisfy all of them, written: $\varphi_n \doteq \E_{2n+1} \limp (\varphi_{request} \land \varphi^{true}_{2n+1})$.

It should be noted the following: \emph{in \requests benchmarks, the size of the progressed \LTLf goals may grow exponentially wrt that of the non-progressed goals}. To see this, observe the following: at each time step, when the environment issues a specific request, the \LTLf progression of the head of the implication in $\varphi_{request}$ has exponential size due to the nesting of $\meventually$, $\mnext$ and $\land$. 

In fact, we observe that we have an analogous result in \plants and \tireworld benchmarks as well.

\section{Additional Empirical Results}\label{sec:experiments_bis}

In this section, we give additional empirical results, specific for each instance in our benchmarks. For convenience, we present the empirical results grouped by benchmark. Also, we remark that the figures in this section are in \emph{log scale} for better readability.

\smallskip

\begin{figure*}[t]
\centering
\begin{subfigure}{0.365\linewidth}
\includegraphics[width=\linewidth]{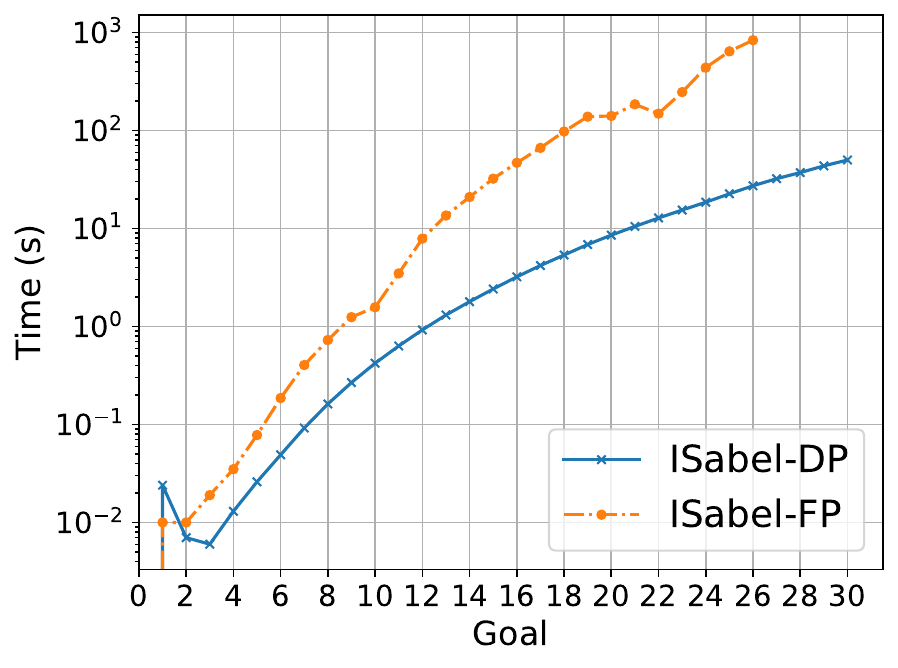}
\end{subfigure}
\begin{subfigure}{0.365\linewidth}
\includegraphics[width=\linewidth]{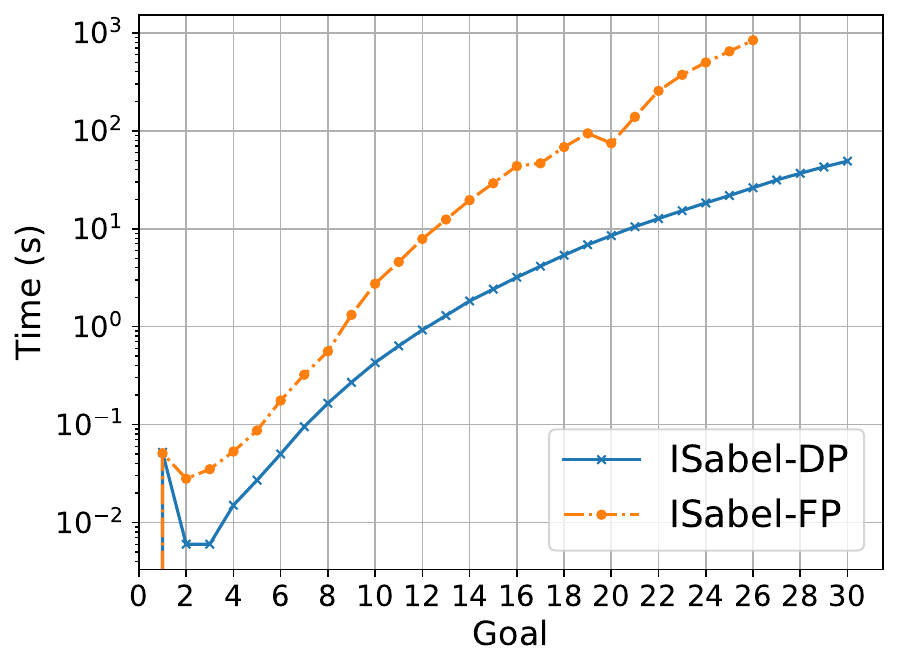}
\end{subfigure}
\begin{subfigure}{0.365\linewidth}
\includegraphics[width=\linewidth]{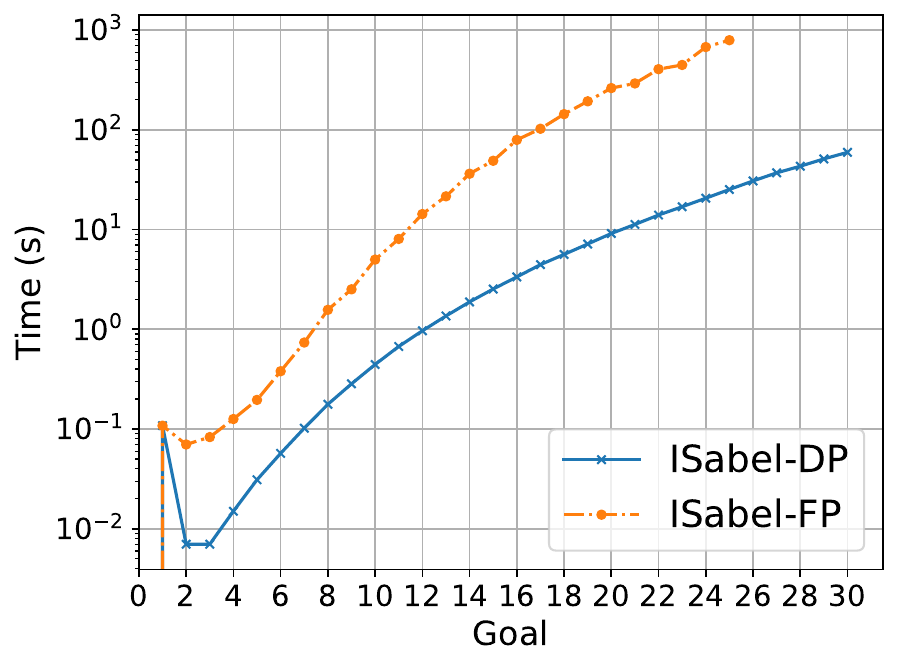}
\end{subfigure}
\begin{subfigure}{0.365\linewidth}
\includegraphics[width=\linewidth]{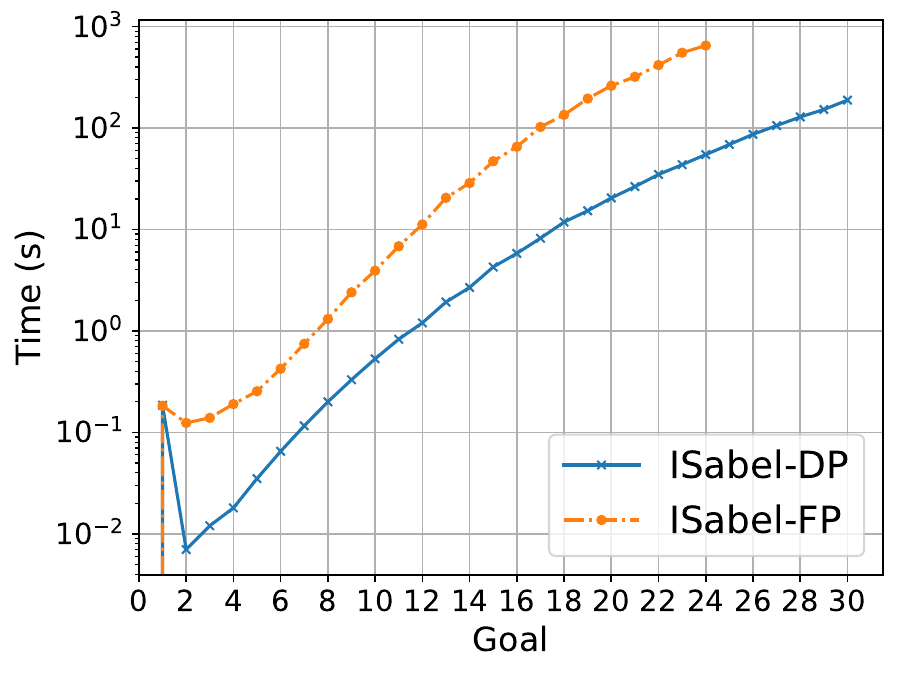}
\end{subfigure}
\begin{subfigure}{0.365\linewidth}
\includegraphics[width=\linewidth]{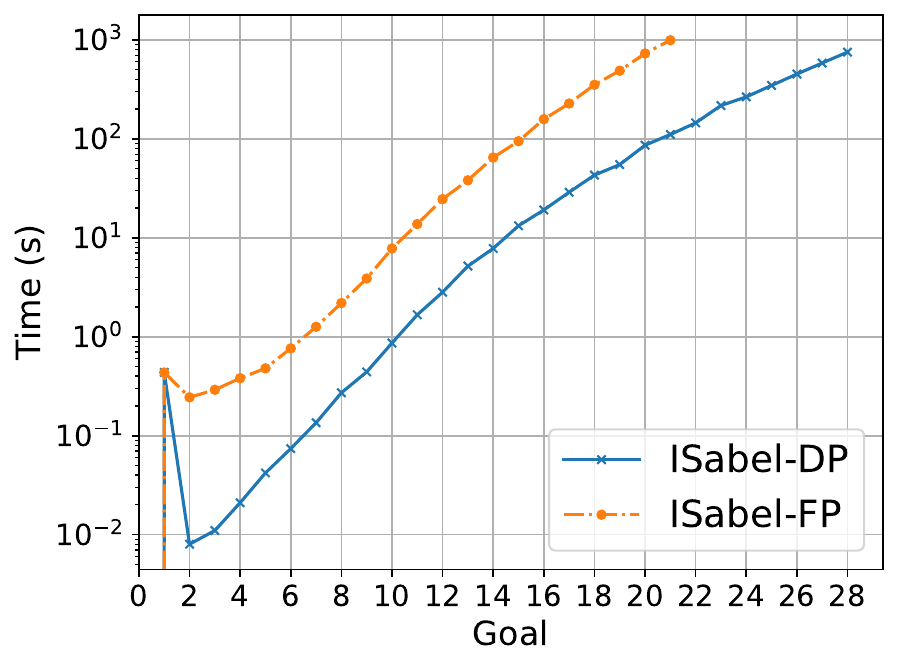}
\end{subfigure}
\begin{subfigure}{0.365\linewidth}
\includegraphics[width=\linewidth]{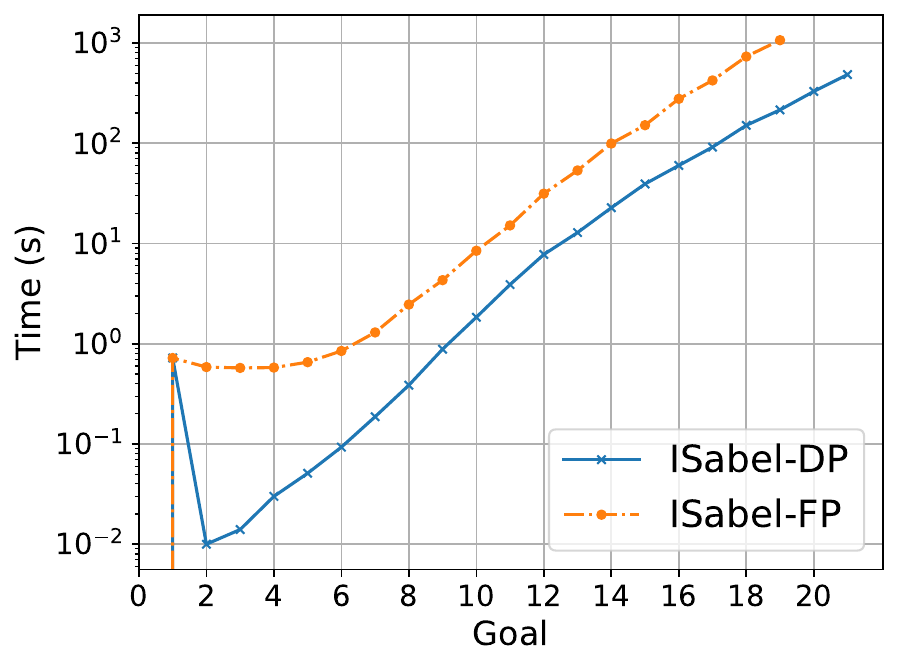}
\end{subfigure}
\begin{subfigure}{0.365\linewidth}
\includegraphics[width=\linewidth]{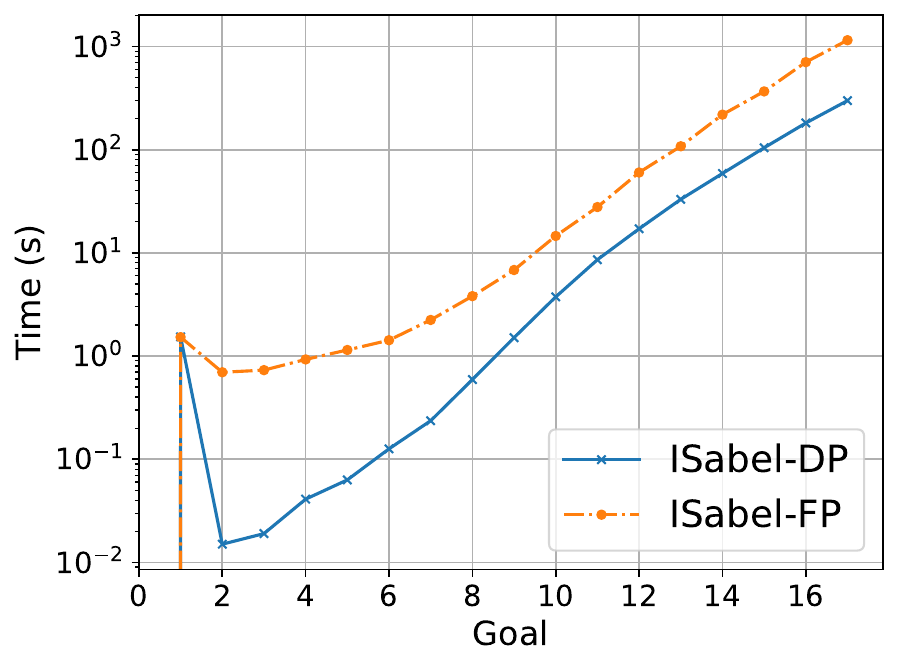}
\end{subfigure}
\begin{subfigure}{0.365\linewidth}
\includegraphics[width=\linewidth]{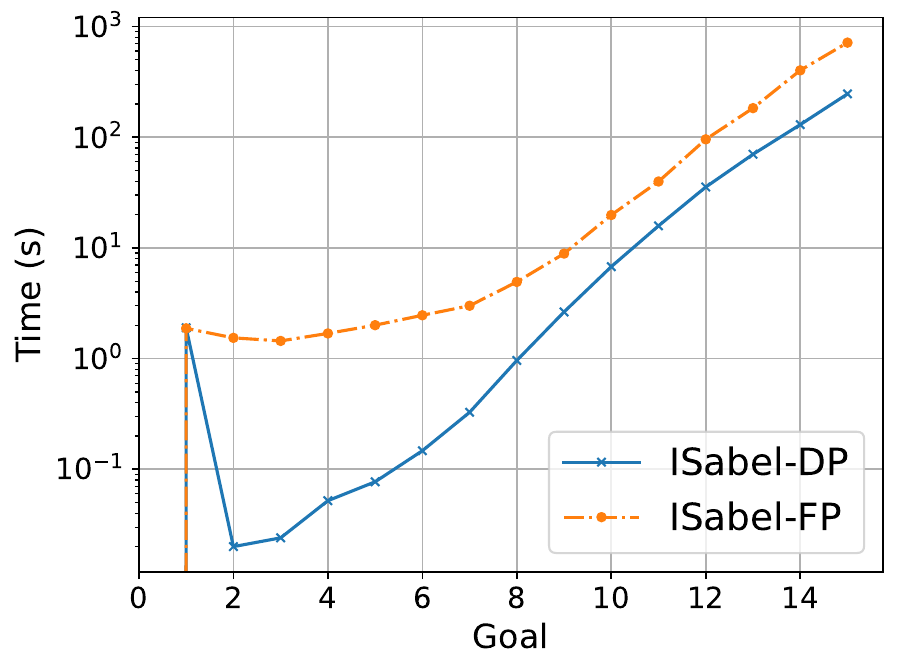}
\end{subfigure}
\begin{subfigure}{0.365\linewidth}
\includegraphics[width=\linewidth]{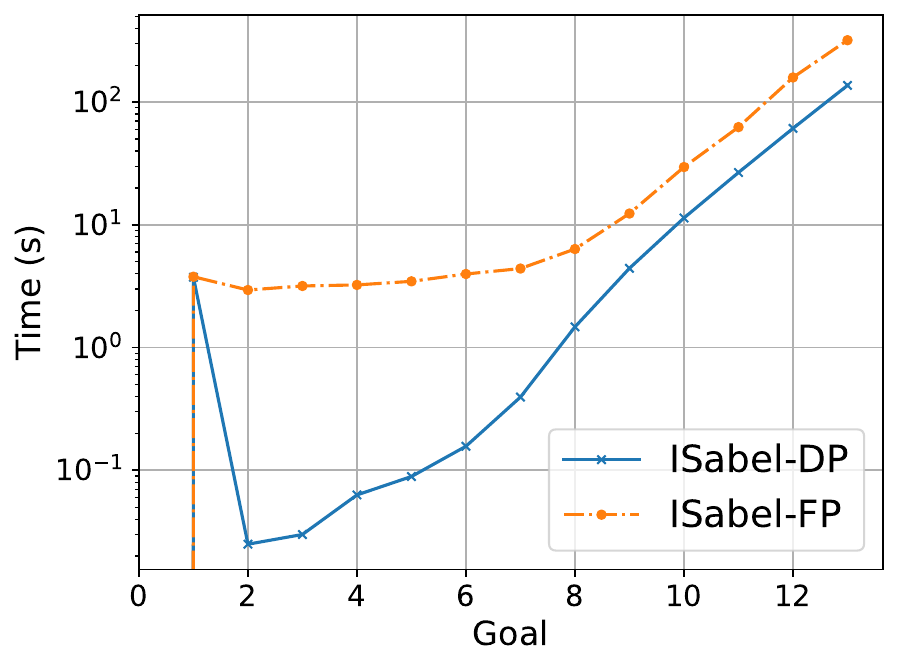}
\end{subfigure}
\begin{subfigure}{0.365\linewidth}
\includegraphics[width=\linewidth]{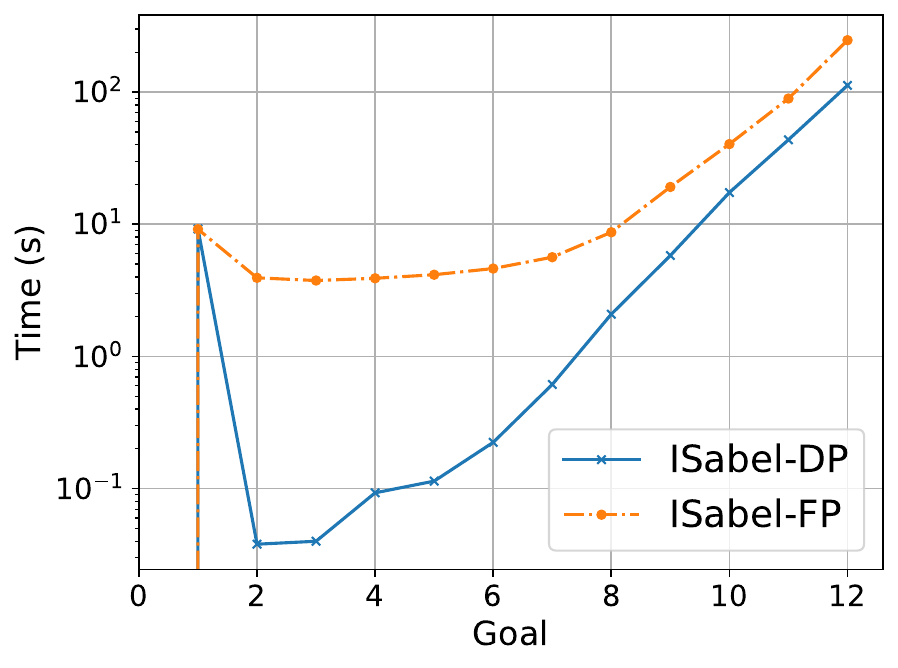}
\end{subfigure}
\caption{From top to bottom, left to right, time required by \isabeldp and \isabelfp (in \emph{log} scale) to add new goals in instances \tireworld-$1$ to \tireworld-$10$.}
\label{fig:tireworld}
\end{figure*}

\noindent \textbf{\tireworld.} Figure~\ref{fig:tireworld} compares the time required by \isabeldp and \isabelfp to add a goal in each \tireworld instance. We observe that \isabeldp requires less time to add a new goal than \isabelfp in every \tireworld instance -- often gaining one or two orders of magnitude better performance. In fact, the performance gap between \isabeldp and \isabelfp becomes smaller as the number of locations $\ell$ increases; this is due to the memory management overhead that we discussed in Section~\ref{sec:experiments} of the body of the paper. Nonetheless, the better \DFA game construction technique employed by \isabeldp suggests that, on more powerful hardware, it scales better than \isabelfp in large \tireworld instances as well. 

Regarding \isabeldp, observe that the time required to add the first (aka original) \LTLf goal -- which also includes the \LTLf domain specification -- is orders of magnitude higher than that required to add some of the subsequent \LTLf goals. The reason is that the size of the original goal is dominated by the size of the conjuncts that constitute the domain specification; the subsequent added goals do not include the domain specification, meaning that they often have size much smaller than that of the original goal. The performance gap between the time required by \isabeldp to add the original goal and that to add some of the subsequent goals is then explained since the \DFAs of the subsequent goals can be constructed efficiently compared with the \DFA of the original goal. Indeed, the performance gap is more evident in large \tireworld instances -- where the size of the domain specification is larger.

Note that \isabelfp does not exhibit the same behavior: the time required to add the original goal is in the same order of magnitude of the time required to add some of the subsequent goals. Indeed, to add a new goal, \isabelfp constructs the \DFA of the \LTLf specification obtained as the conjunction of the progressed \LTLf goals -- i.e., every time \isabelfp adds a new goal, it constructs from scratch the \DFA of part of the \LTLf domain specification. As a result, even if some of the subsequent goals are orders of magnitude smaller than the original goal, \isabelfp incurs the unnecessary overhead of constructing again the \DFA of the domain specification.

\smallskip

\begin{figure*}[t]
\centering
\begin{subfigure}{0.365\linewidth}
\includegraphics[width=\linewidth]{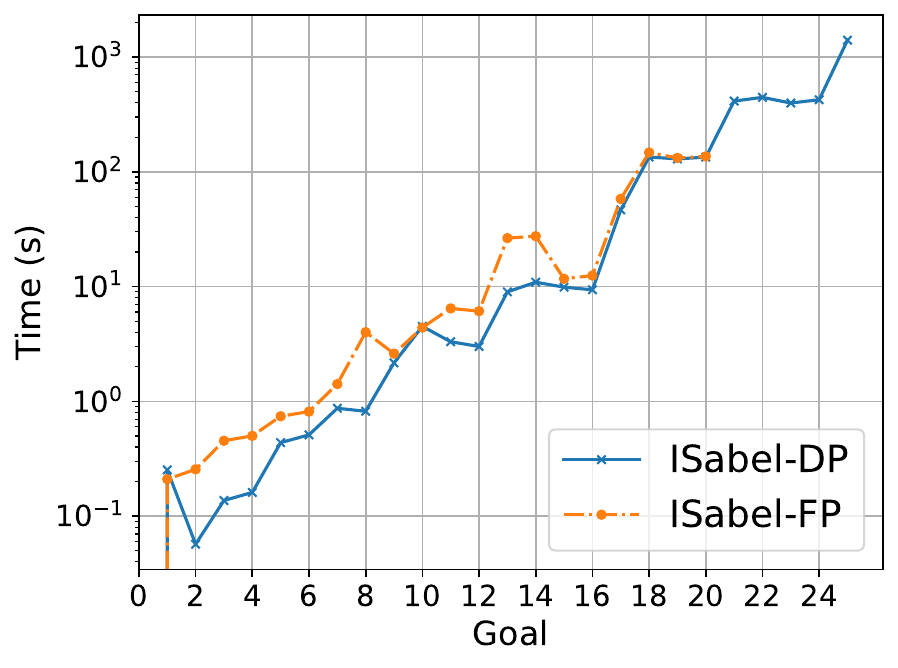}
\end{subfigure}
\begin{subfigure}{0.365\linewidth}
\includegraphics[width=\linewidth]{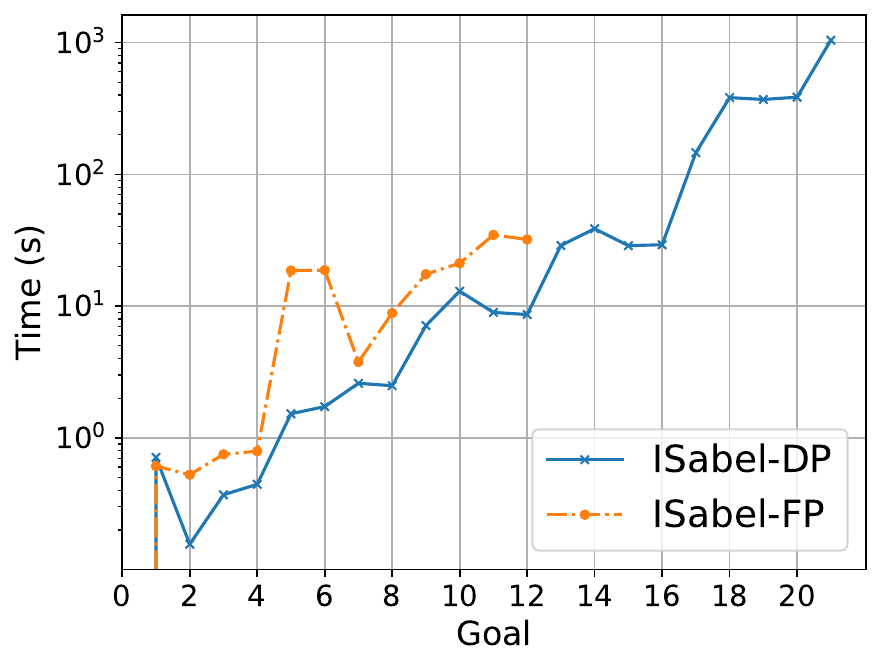}
\end{subfigure}
\begin{subfigure}{0.365\linewidth}
\includegraphics[width=\linewidth]{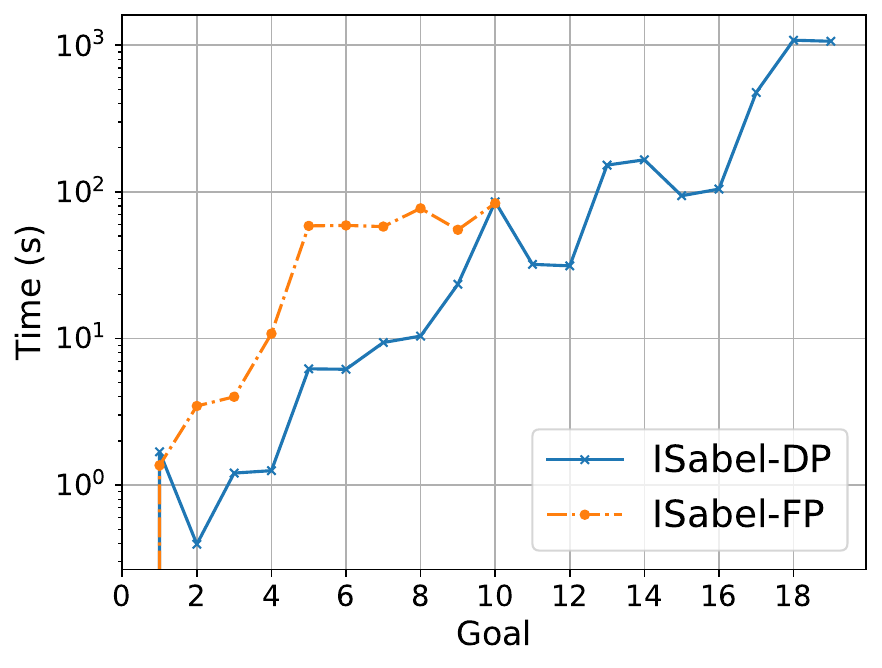}
\end{subfigure}
\caption{From top to bottom, left to right, time required by \isabeldp and \isabelfp (in \emph{log} scale) to add new goals in instances \counter-$6$ to \counter-$8$.}
\label{fig:counter}
\end{figure*}

\noindent \textbf{\counter.} Figure~\ref{fig:counter} compares the time required by \isabeldp and \isabelfp to add a goal in each \counter instance. We observe that \isabeldp is faster than \isabelfp to add goals in every \counter instance. Furthermore, we note that the performance gap between the time required by \isabeldp to add the original goal and that to add some of the subsequent goals can be observed in \counter instances as well, even through it is less pronounced compared to \tireworld instances. This latter result, together with the fact that in \counter instances the size of the progressed \LTLf goals is almost the same as that of the non-progressed goals, see Section~\ref{sec:specs}, gives that the performance gap between \isabeldp and \isabelfp is mostly explained since \isabeldp benefits from storing in memory the \DFAs of the previously added \LTLf goals.

\smallskip 

\begin{figure*}[t]
\centering
\begin{subfigure}{0.365\linewidth}
\includegraphics[width=\linewidth]{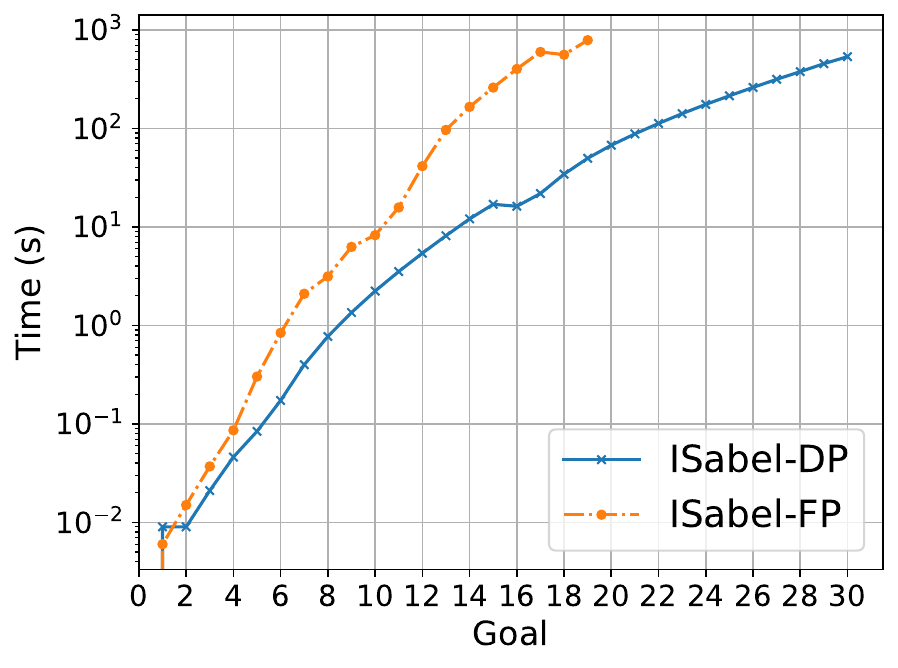}
\end{subfigure}
\begin{subfigure}{0.365\linewidth}
\includegraphics[width=\linewidth]{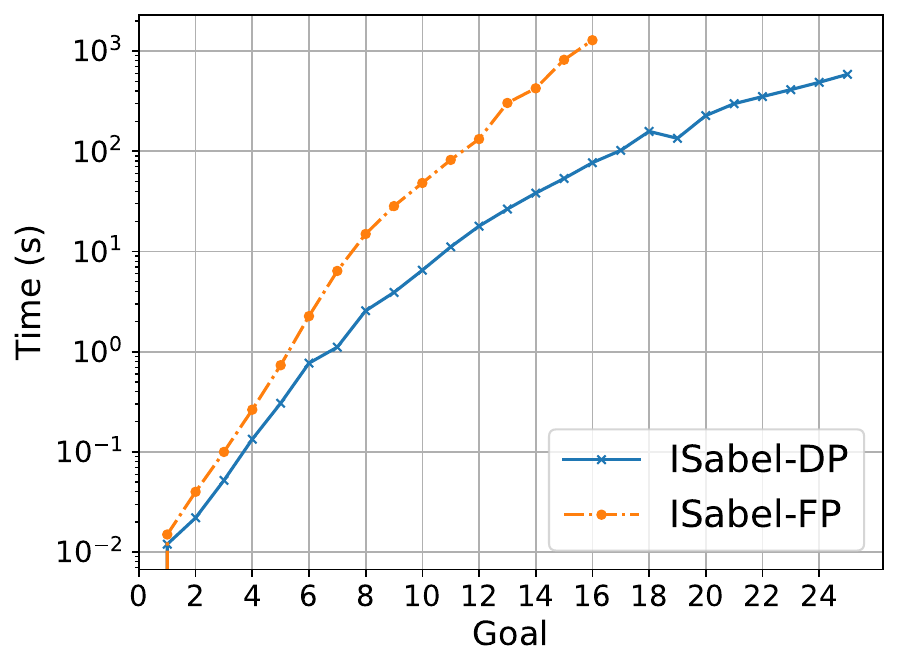}
\end{subfigure}
\begin{subfigure}{0.365\linewidth}
\includegraphics[width=\linewidth]{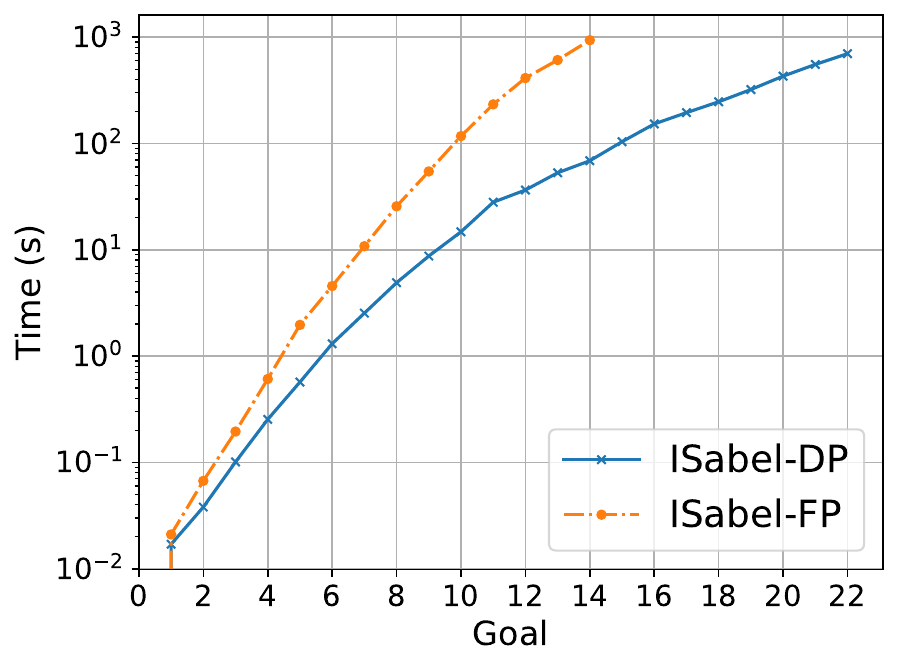}
\end{subfigure}
\begin{subfigure}{0.365\linewidth}
\includegraphics[width=\linewidth]{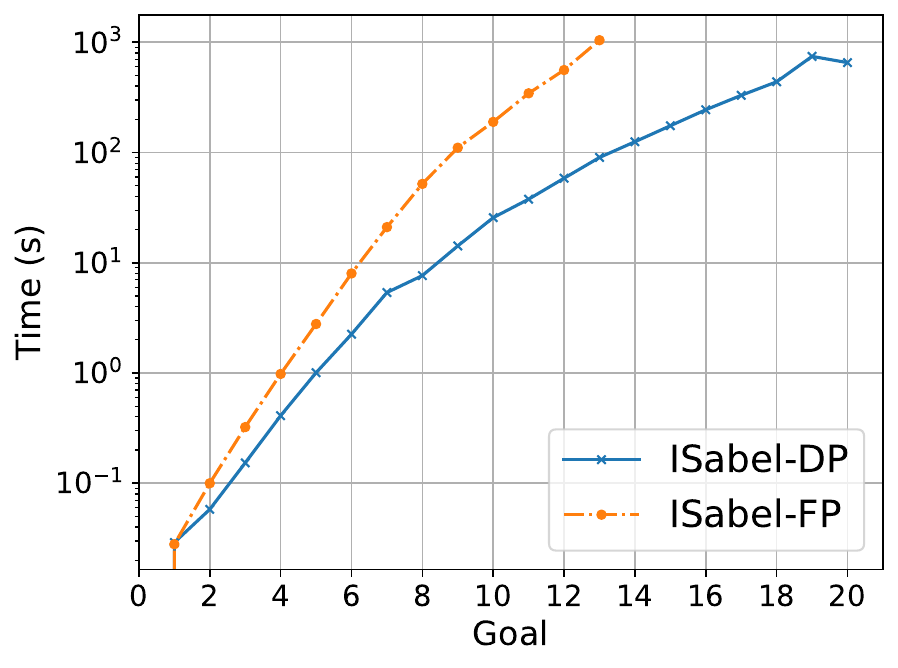}
\end{subfigure}
\begin{subfigure}{0.365\linewidth}
\includegraphics[width=\linewidth]{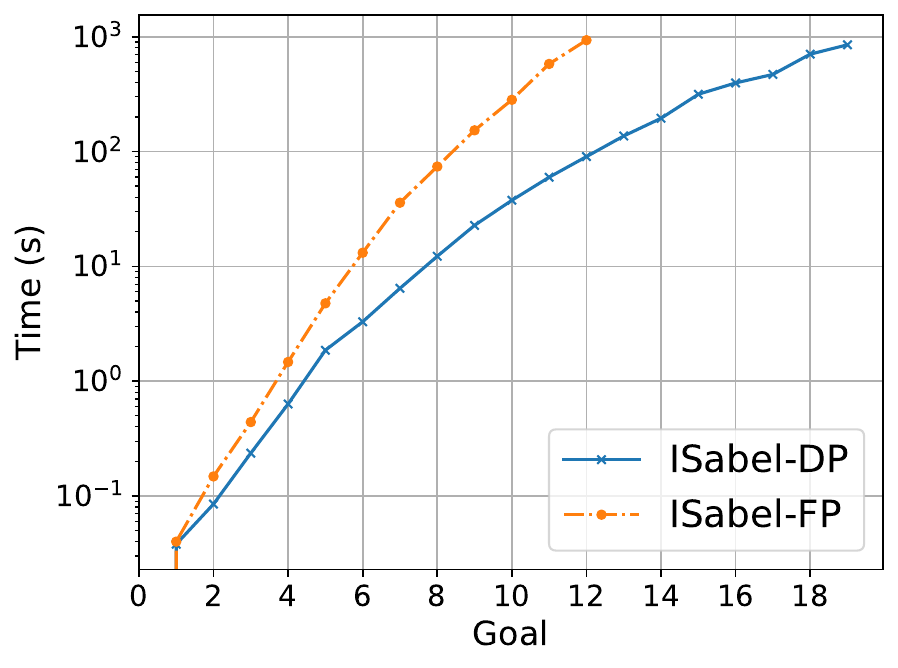}
\end{subfigure}
\begin{subfigure}{0.365\linewidth}
\includegraphics[width=\linewidth]{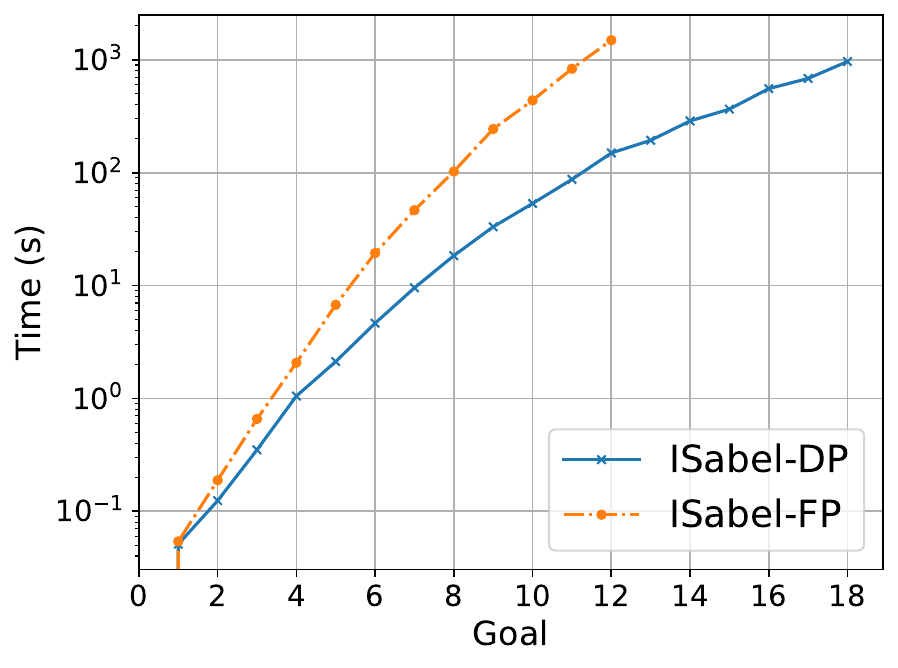}
\end{subfigure}
\begin{subfigure}{0.365\linewidth}
\includegraphics[width=\linewidth]{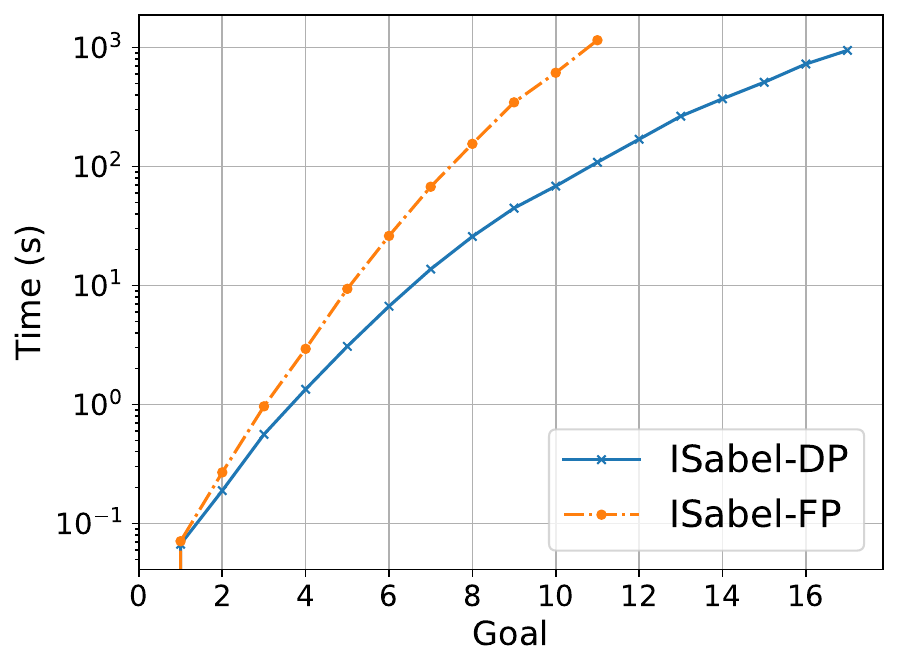}
\end{subfigure}
\begin{subfigure}{0.365\linewidth}
\includegraphics[width=\linewidth]{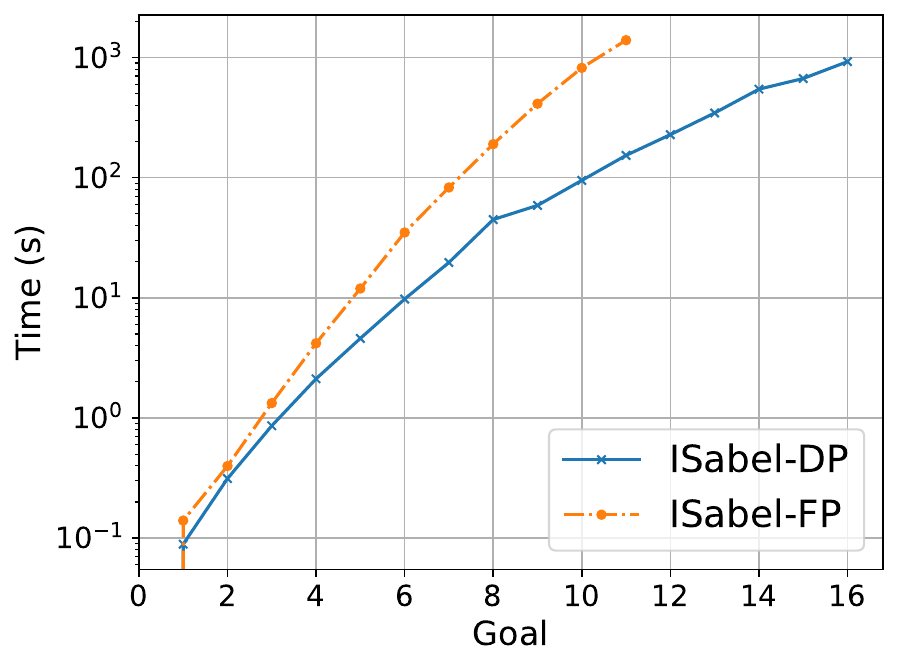}
\end{subfigure}
\begin{subfigure}{0.365\linewidth}
\includegraphics[width=\linewidth]{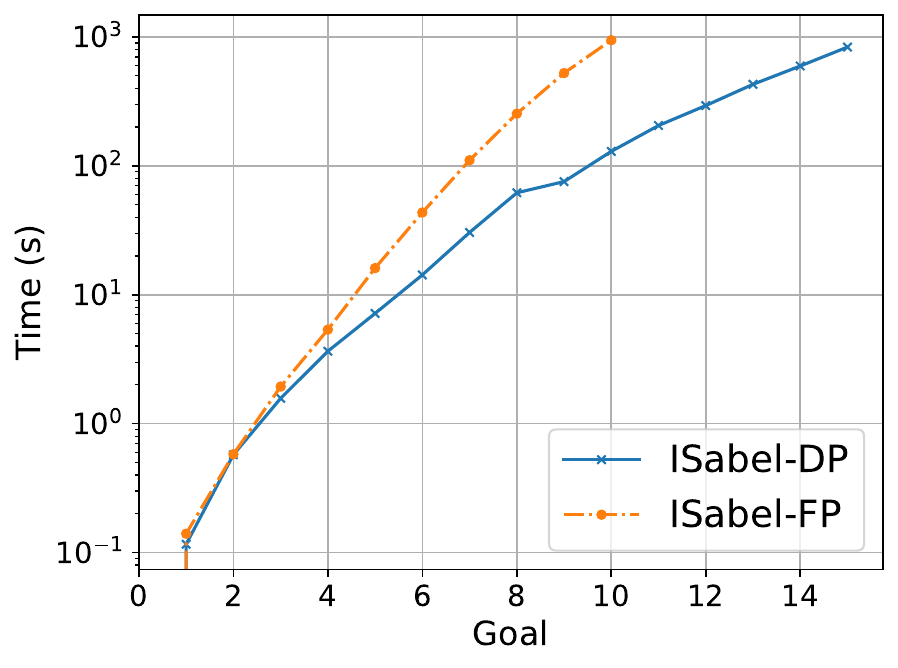}
\end{subfigure}
\begin{subfigure}{0.365\linewidth}
\includegraphics[width=\linewidth]{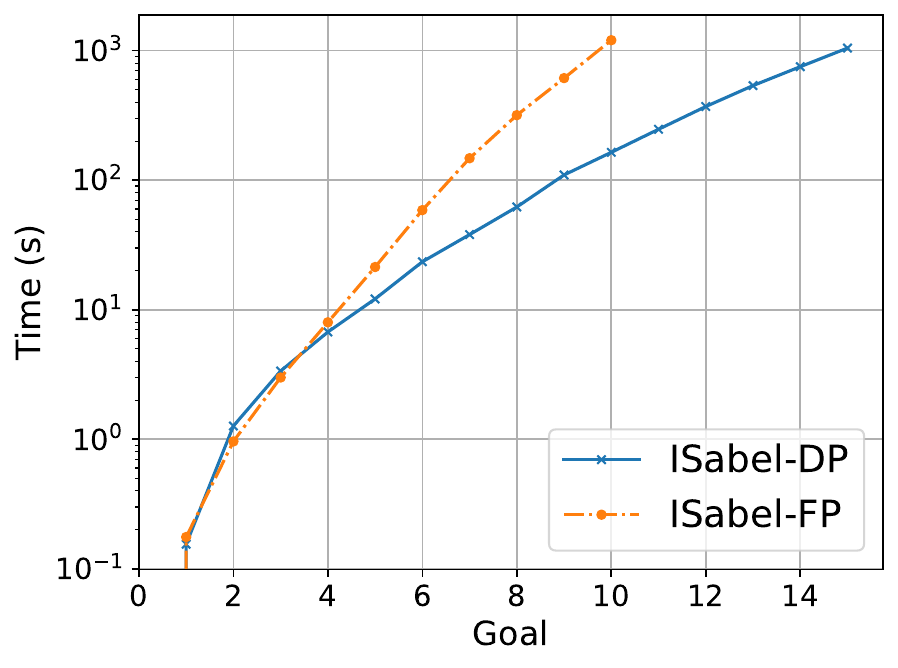}
\end{subfigure}
\caption{From top to bottom, left to right, time required by \isabeldp and \isabelfp (in \emph{log} scale) to add new goals in instances \plants-$1$ to \plants-$10$.}
\label{fig:plants}
\end{figure*}

\begin{figure*}[t]
\centering
\begin{subfigure}{0.33\linewidth}
\includegraphics[width=\linewidth]{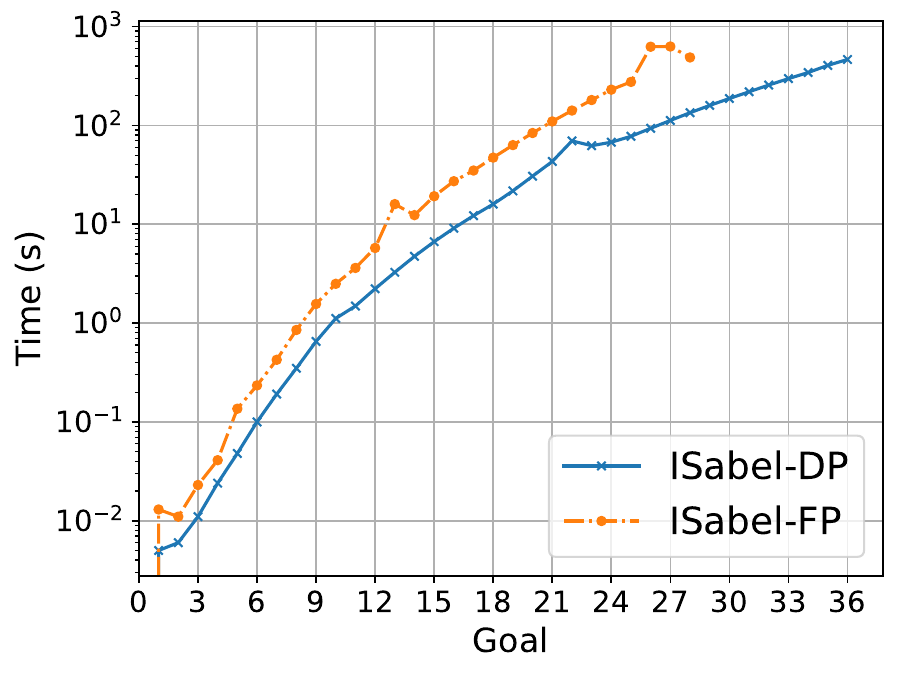}
\end{subfigure}
\begin{subfigure}{0.33\linewidth}
\includegraphics[width=\linewidth]{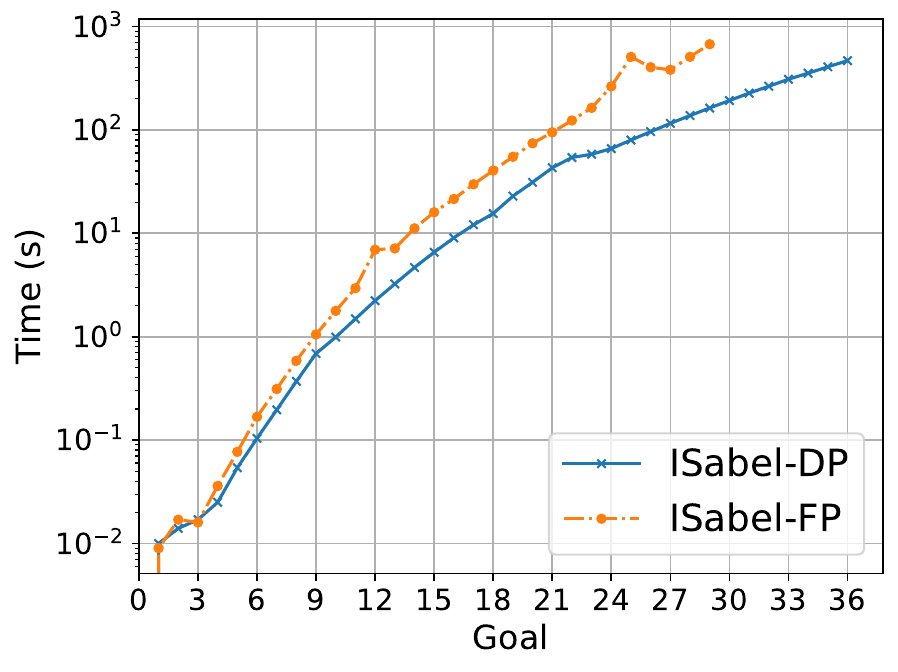}
\end{subfigure}
\begin{subfigure}{0.33\linewidth}
\includegraphics[width=\linewidth]{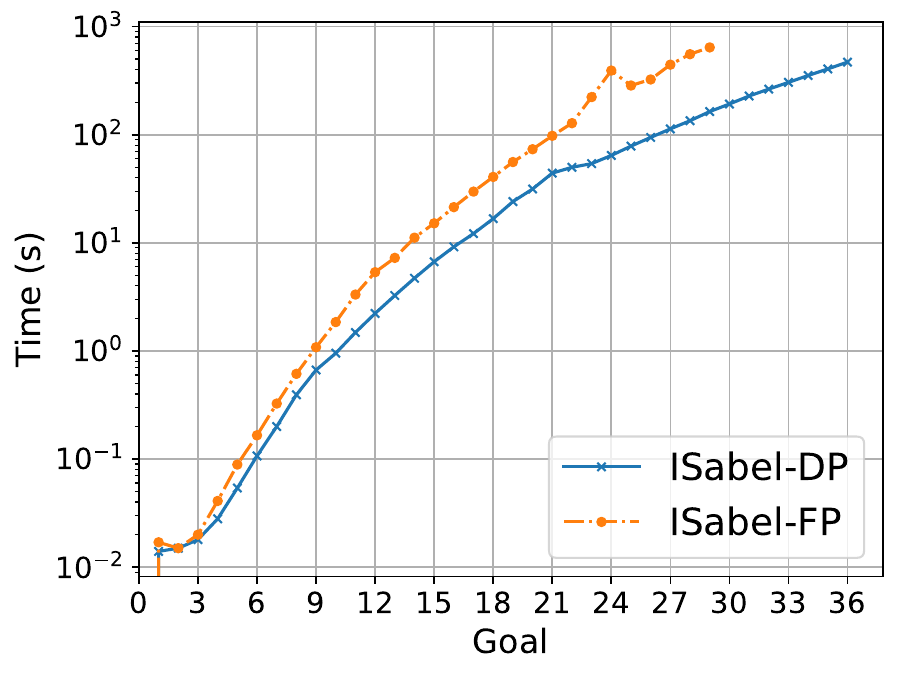}
\end{subfigure}
\begin{subfigure}{0.33\linewidth}
\includegraphics[width=\linewidth]{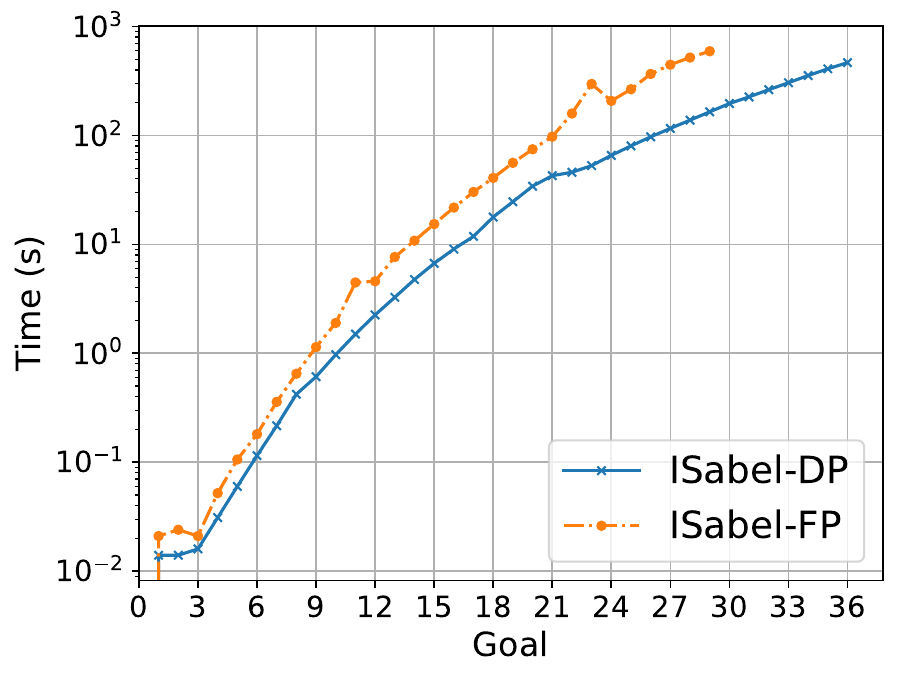}
\end{subfigure}
\begin{subfigure}{0.33\linewidth}
\includegraphics[width=\linewidth]{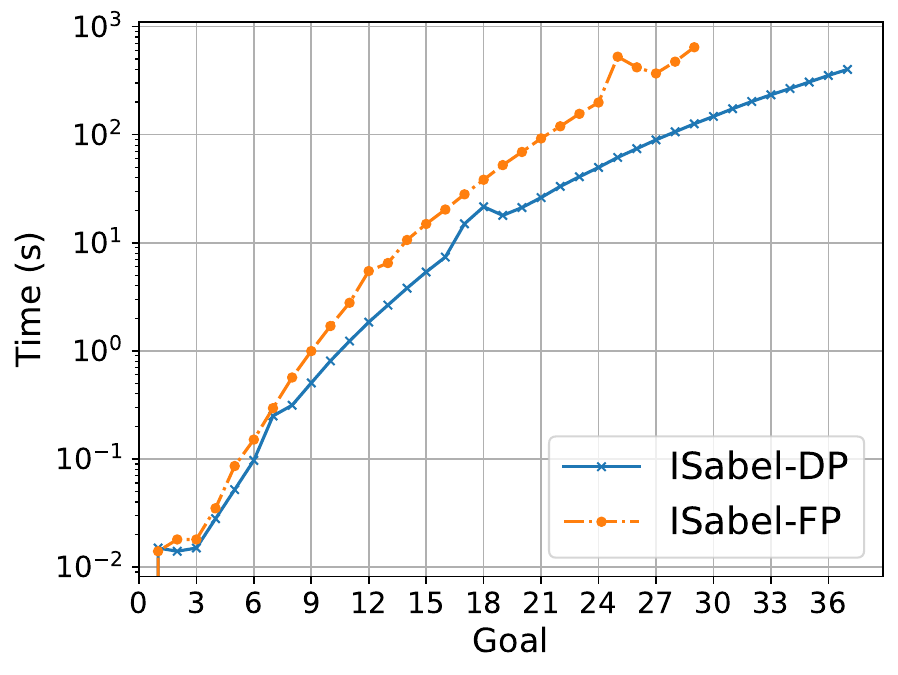}
\end{subfigure}
\begin{subfigure}{0.33\linewidth}
\includegraphics[width=\linewidth]{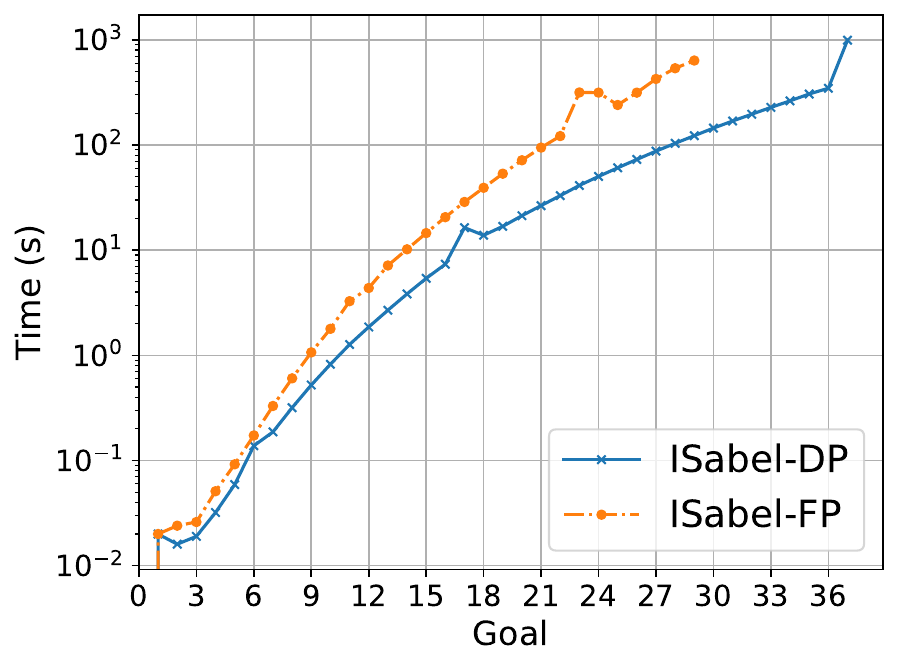}
\end{subfigure}
\begin{subfigure}{0.33\linewidth}
\includegraphics[width=\linewidth]{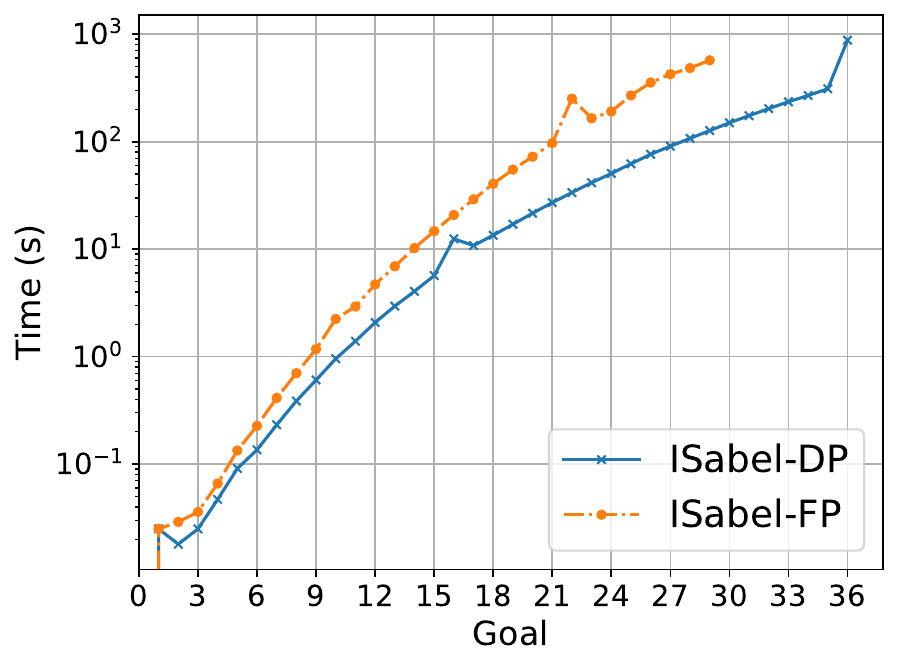}
\end{subfigure}
\begin{subfigure}{0.33\linewidth}
\includegraphics[width=\linewidth]{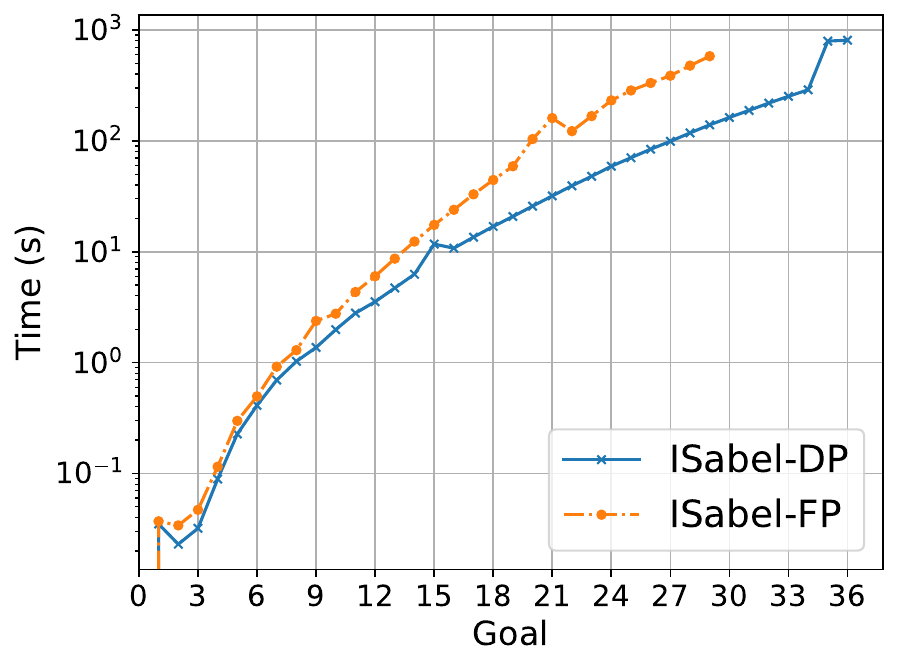}
\end{subfigure}
\begin{subfigure}{0.33\linewidth}
\includegraphics[width=\linewidth]{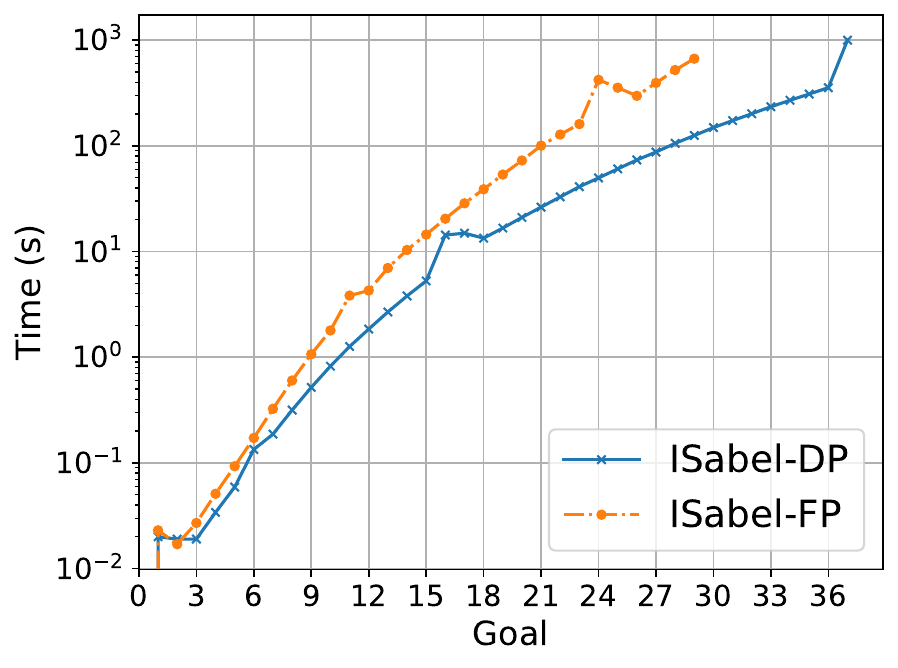}
\end{subfigure}
\begin{subfigure}{0.33\linewidth}
\includegraphics[width=\linewidth]{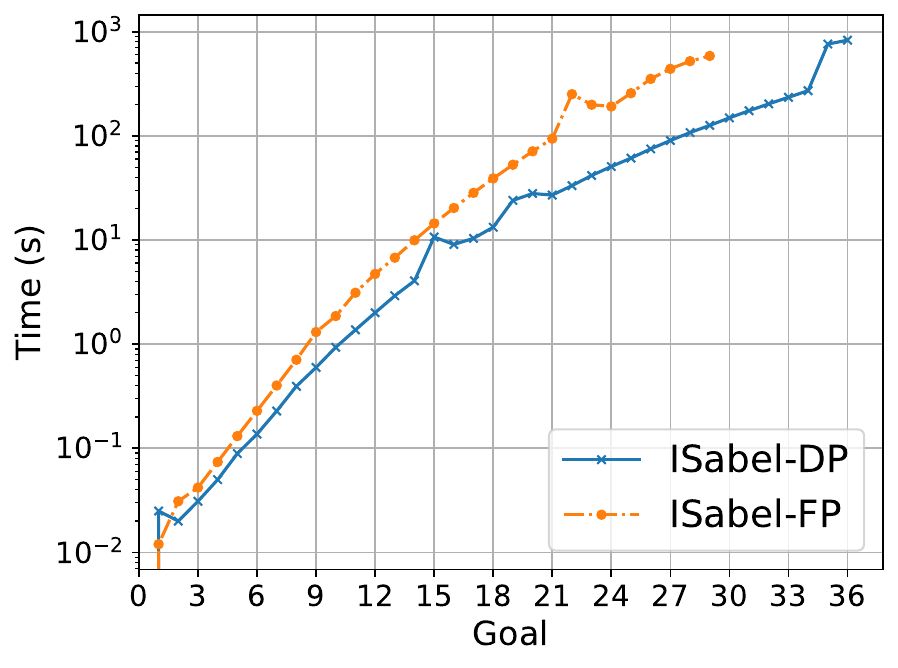}
\end{subfigure}
\begin{subfigure}{0.33\linewidth}
\includegraphics[width=\linewidth]{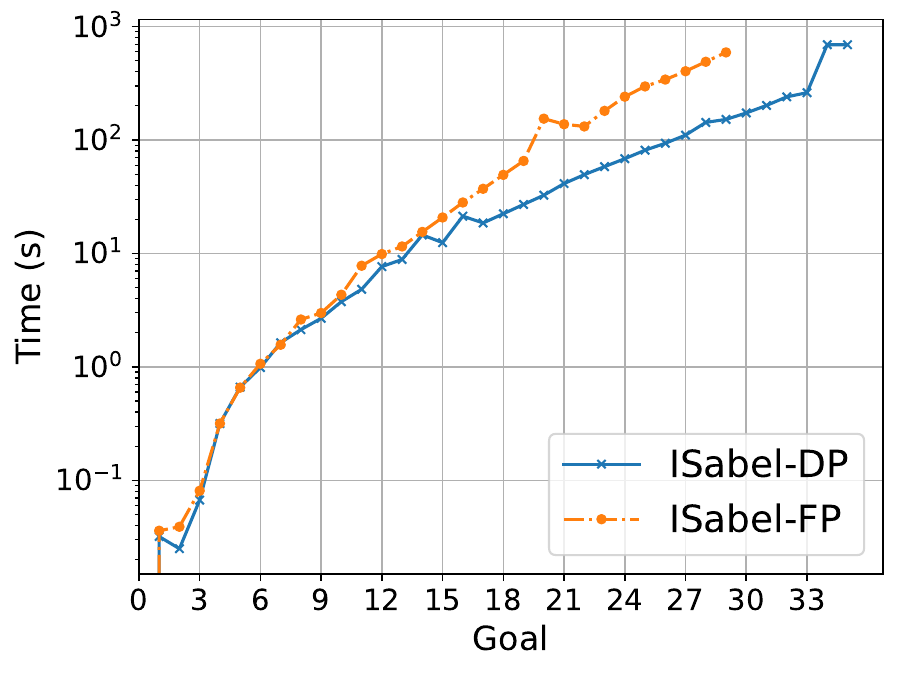}
\end{subfigure}
\begin{subfigure}{0.33\linewidth}
\includegraphics[width=\linewidth]{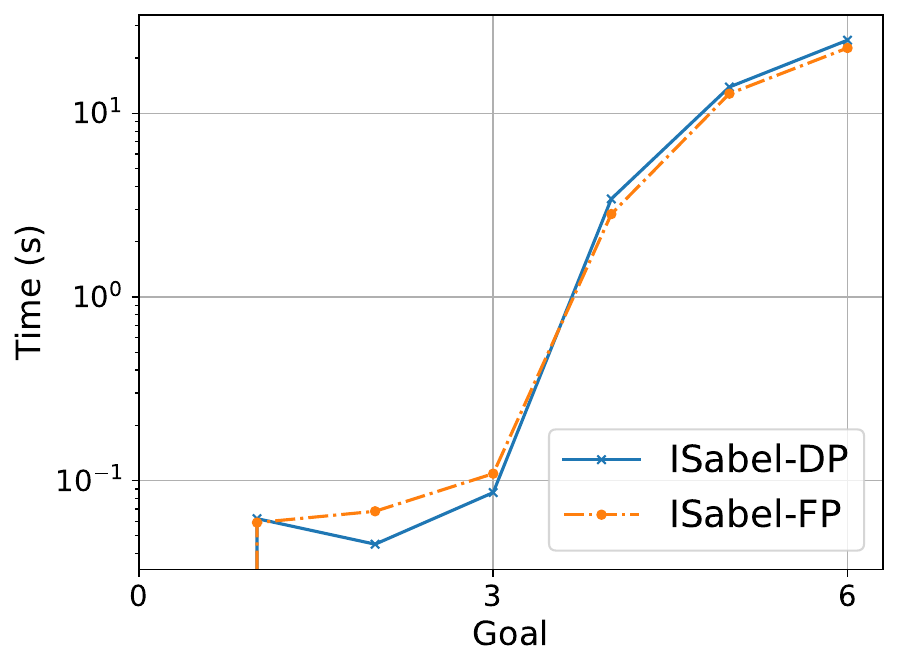}
\end{subfigure}
\caption{From top to bottom, left to right, time required by \isabeldp and \isabelfp (in \emph{log} scale) to add new goals in instances \requests-$1$-$1$ to \requests-$3$-$4$.}
\label{fig:requests}
\end{figure*}

\noindent \textbf{\plants} and \textbf{\requests}. Figures~\ref{fig:plants}
and~\ref{fig:requests} compare the time required by \isabeldp and \isabelfp to add a goal in each \plants and \requests instance, respectively. The results show that, when the number of goals to be added is small, \isabeldp and \isabelfp may be comparable; however, if many goals need to be added during execution, then \isabeldp is clearly superior (with the obvious exception of \requests-$3$-$4$, where both \isabeldp and \isabelfp time out while constructing the \DFA of the $7$-th goal). This is due to the following: in the earliest time steps the size of the progressed \LTLf goals does not blowup -- meaning that \isabeldp and \isabelfp have comparable performances; however, after a certain number of time steps the size of the progressed \LTLf goals increases considerably, and hence the overhead \isabelfp incurs becomes relevant -- which causes \isabeldp to scale better than \isabelfp when many goals are added.

\end{document}